%% file: main.tex
\documentclass[fleqn,10pt]{wlscirep}
\usepackage[utf8]{inputenc}
\usepackage[T1]{fontenc}

\usepackage{subfig}
\usepackage{graphicx}

\usepackage{xcolor, soul}
\sethlcolor{white}
\newcommand{\mathcolorbox}[1]{\colorbox{white}{$\displaystyle #1$}}

\usepackage{multicol}

\usepackage{wrapfig}
\usepackage{lscape}
\usepackage{rotating}
\usepackage{graphicx}
\usepackage{caption}
\usepackage{amsmath}
\usepackage{bm}
\usepackage{mathtools}
\usepackage{multirow,booktabs}
\usepackage{siunitx}
\usepackage{adjustbox}
\usepackage{graphicx}
\usepackage{float}
\usepackage{marginnote}
\usepackage{longtable}
\usepackage{supertabular}
\usepackage{breqn}

\usepackage{titlesec}

\graphicspath{{Pictures/}} 
\usepackage[square, numbers]{natbib} 

\makeatletter
\newcommand{\customlabel}[2]{%
\protected@write \@auxout {}{\string \newlabel {#1}{{#2}{}}}}
\makeatother

\usepackage{hyperref}

\makeatletter
\def\ttabular{%
\hbox\bgroup
\let\\\cr
\def\rulea{\ifnum\rowc=\@ne \hrule height 1.3pt \fi}
\def\ruleb{
\ifnum\rowc=1\hrule height 1.3pt \else
\ifnum\rowc=6\hrule height \heavyrulewidth 
   \else \hrule height \lightrulewidth\fi\fi}
\valign\bgroup
\global\rowc\@ne
\rulea
\hbox to 10em{\strut \hfill##\hfill}%
\ruleb
&&%
\global\advance\rowc\@ne
\hbox to 10em{\strut\hfill##\hfill}%
\ruleb
\cr}
\def\endttabular{%
\crcr\egroup\egroup}

\title{Dimensional synthesis of spatial manipulators for velocity and force transmission for operation around a specified task point}

\author[1,*]{Akkarapakam Suneesh Jacob}
\author[1,+]{Bhaskar Dasgupta}

\affil[1]{Affiliation: Indian Institute of Technology Kanpur, Kanpur, India}

\affil[*]{Emails: sunjac@iitk.ac.in, suneeshjacob@gmail.com}
\affil[+]{Email: dasgupta@iitk.ac.in}

\keywords{Enumeration, Structural synthesis, Mechanisms, Spatial manipulators}

\begin{abstract}
Dimensional synthesis refers to design of the dimensions of manipulators by optimising different kinds of performance indices. The motivation of this study is to perform dimensional synthesis for a wide set of spatial manipulators by optimising the manipulability of each manipulator around a pre-defined task point in the workspace and to finally give a prescription of manipulators along with their dimensions optimised for velocity and force transmission. A systematic method to formulate Jacobian matrix of a manipulator is presented. Optimisation of manipulability is performed for manipulation of the end-effector around a chosen task point for 96 1-DOF manipulators, 645 2-DOF manipulators, 8 3-DOF manipulators and 15 4-DOF manipulators taken from the result of enumeration of manipulators that is done in its companion paper devoted to enumeration of possible manipulators up to a number of links. Prescriptions for these sets of manipulators are presented along with their scaled condition numbers and their ordered indices. This gives the designer a prescription of manipulators with their optimised dimensions that reflects the performance of the end-effector around the given task point for velocity and force transmission.\newline

\emph{Keywords:} dimensional synthesis, manipulability, condition number, jacobian, optimisation

\end{abstract}

\begin{document}
\twocolumn
\flushbottom

\maketitle
%
%
\thispagestyle{empty}


\section{Introduction}

In the context of this study, a manipulator is defined as an assembly of kinematic links connected through kinematic joints, with a fixed base link and a moving end-effector link, in which the end-effector link’s motion and force-transformation are controlled by actuating one or more kinematic joints. A kinematic link is a rigid link that can transfer motion and forces, and a kinematic joint is a connection between two links that allows relative motion between the two links. Dimensional synthesis of a manipulator is the problem of designing an appropriate set of dimensions for the manipulator of defined topology\footnote{A specific set of connections of links with specific types of joints} to achieve a requirement. Dimensional synthesis problems are generally solved by taking one or more objectives into consideration, such as energy optimisation and well-conditioning of the end-effector's motion. This study focuses on the objective of well-conditioning for velocity and force transmission around a given task point. Well-conditioning is a measure of being distant from singularity. Singularity is related to the local degeneracy of input-output motion/force transformation. A manipulator is said to have a singularity at a point in the joint space if at that configuration the end-effector of the manipulator is unable to be controlled by the actuating joint velocities at least in one direction, either in translation or by rotation. The singularity of a manipulator is often analysed using its Jacobian, the transformation matrix of the joint velocities to the velocity of the end-effector. Using the principle of virtual work, it can be shown that for a system with negligible frictional power losses the force transmission matrix is exactly the transpose of the velocity transmission matrix. This gives an advantage in designing the dimensions of the manipulators for both inverse force transformation and direct velocity transformation by using the same matrix. Hence, this matrix is used to analyse the singularities of the manipulator. The task of dimensional synthesis is to design the dimensional parameters of various manipulators with the criterion of the maximum extent of singularity avoidance when a representative point in the workspace about which the manipulator's end-effector operates is given along with the bounds of the environment. The criterion used in this study is the manipulability index. The manipulability index for a manipulator would typically be a function of dimensional parameters as variables. The manipulability index with ellipsoidal approach \cite{yoshikawa1985manipulability} is as follows.

\begin{align}
    \mu=\left\{
                \begin{array}{ll}
                  \sqrt{\det{\left([J][J]^{T}\right)}} \> \> \text{if} \> \> n_j > n_t \\
                  \sqrt{\det{\left([J]^{T}[J]\right)}} \> \> \text{if} \> \> n_j \leq n_t
                \end{array}
              \right.
 \label{eq:1}
 \end{align}

\noindent where $n_j$ is the dimension of the joint space and $n_t$ is the dimension of the task space of the manipulator.

The values of the dimensional parameters that would maximise this function would describe the dimensions of the manipulator that would give good performance around the given end-effector point. This concept with ellipsoidal approach is illustrated below for a simple case of a Jacobian matrix that maps linear velocities alone of a 2R-serial planar manipulator.

\begin{figure}[hbt!]
  \centering
  \includegraphics[width=\linewidth]{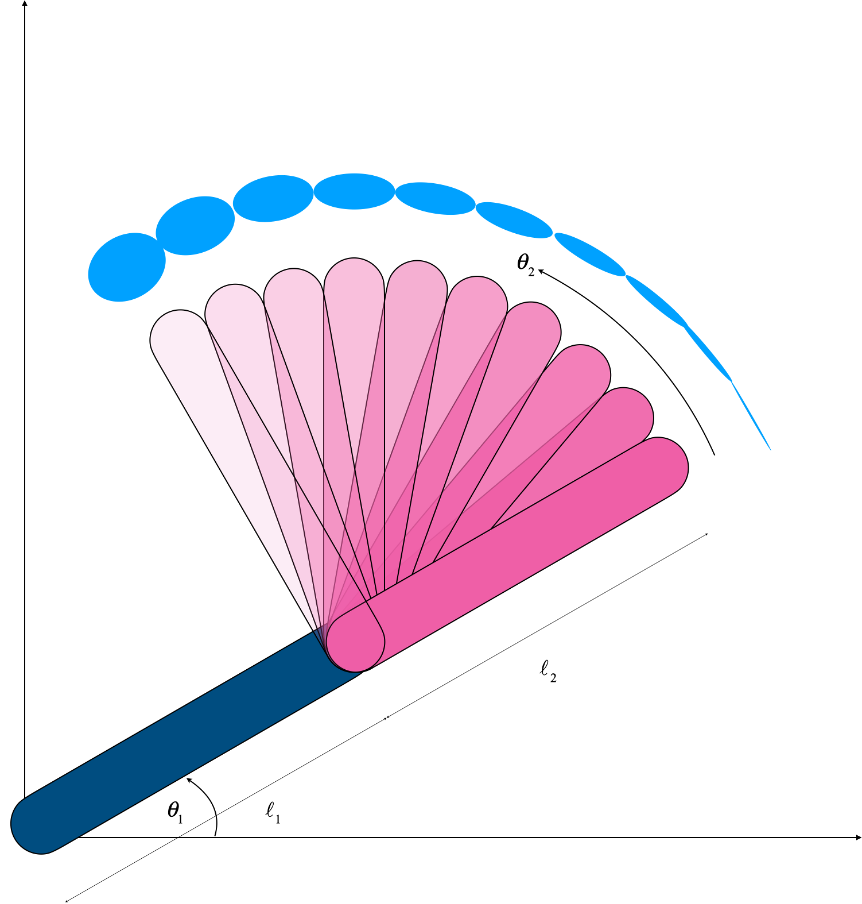}
  \caption{Manipulability varying with $\theta_2$.}
  \label{fig6}
\end{figure}

For a two-link serial manipulator of link lengths $l_1$ and $l_2$ and relative joint angles $\theta_1$ and $\theta_2$ as shown in figure \ref{fig6}, the Jacobian matrix for mapping the linear velocities alone, is given by

\[\begin{Bmatrix}
    v_x \\
    v_y \\
\end{Bmatrix}=
\begin{bmatrix}
    -l_1 \sin{\theta_1}-l_2 \sin{\left(\theta_1+\theta_2\right)}       & -l_2 \sin{\left(\theta_1+\theta_2\right)} \\
    l_1 \cos{\theta_1}+l_2 \cos{\left(\theta_1+\theta_2\right)}       & +l_2 \cos{\left(\theta_1+\theta_2\right)} \\
\end{bmatrix}\begin{Bmatrix}
    \dot{\theta}_1 \\
    \dot{\theta}_2 \\
\end{Bmatrix}
\]
\[
\Rightarrow v_e=[J]\{\dot{\theta}\}
\]

Since the Jacobian matrix in this case is a square matrix, $\sqrt{\det{\left([J]^T[J]\right)}}=\det{([J])}$. The determinant of Jacobian (of transformation of linear velocities) would be

\[\det{\left([J]\right)}=
\begin{vmatrix}
    -l_1 \sin{\theta_1}-l_2 \sin{\left(\theta_1+\theta_2\right)}       & -l_2 \sin{\left(\theta_1+\theta_2\right)} \\
    l_1 \cos{\theta_1}+l_2 \cos{\left(\theta_1+\theta_2\right)}       & +l_2 \cos{\left(\theta_1+\theta_2\right)} \\
\end{vmatrix}
\]
\[
\Rightarrow \det{\left([J]\right)}=l_1l_2\sin{\theta_2}
\]

\begin{figure}[hbt!]
  \centering
  \includegraphics[width=\linewidth]{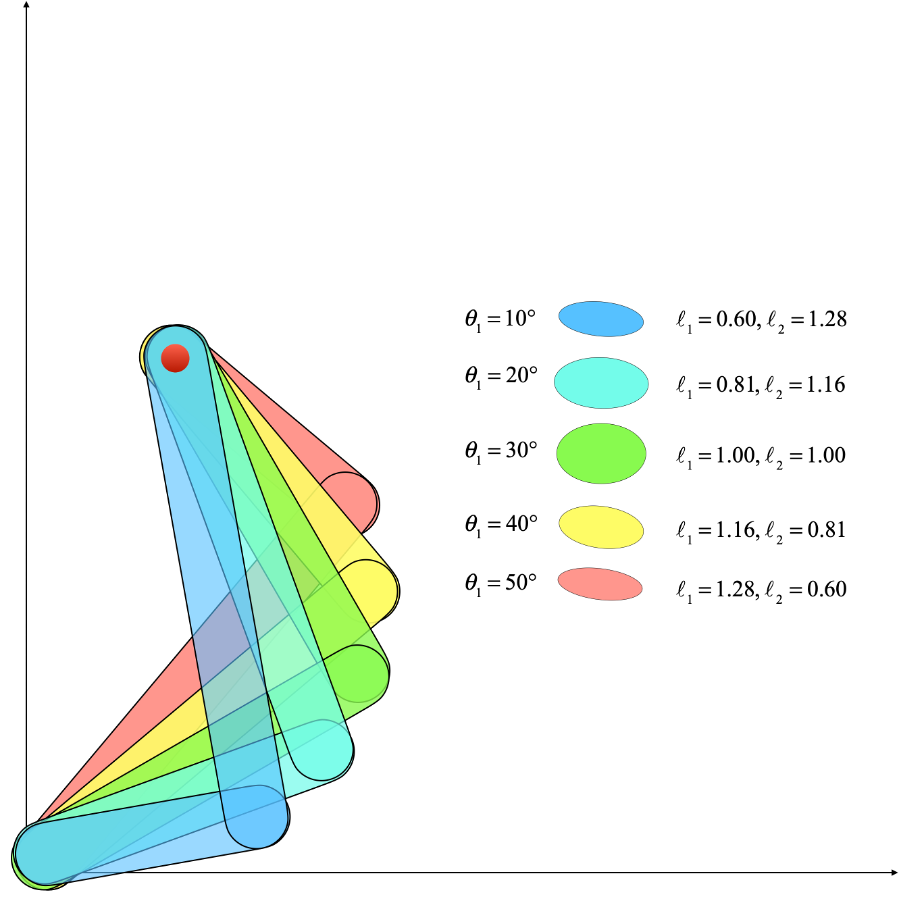}
  \caption{Manipulability varying with $l_1$ and $l_2$.}
  \label{fig7}
\end{figure}

For the determinant to be maximum, the term $l_1l_2 \sin{\theta_2}$ needs to be maximum. Since $l_1$ and $l_2$ are independent of $\theta_2$, by assuming them to be constant, $\theta_2=90^{\circ}$ gives the maximum contribution of manipulability independent of $l_1$ and $l_2$, as shown in the figure \ref{fig6}. And by assuming $l_1$ and $l_2$ to be varying, $l_1=l_2$ gives its maximum contribution, as shown in the figure \ref{fig7}. Hence, for a given location of the end-effector, the values $\theta_1$, $\theta_2$, $l_1$ and $l_2$ can be found.

In the current study, the manipulability index for a specified end-effector point is maximised for various manipulators, and prescriptions are presented for manipulators for same kind of task around the specified point. The same kind of task is reflected by the DOF, and hence the prescriptions of manipulators for each DOF are provided.

\section{Literature review}
Manipulability of a manipulator is calculated by using its Jacobian. In the context of robotics, Jacobian is the matrix that maps the joint-space velocities to the task-space velocities. There are several methods to formulate Jacobian available in the literature. In serial manipulators, only two types of joints, namely prismatic and revolute, are actuated. A standard method to formulate Jacobian for serial manipulators can be found in many books. However, for parallel manipulators, Jacobian formulation becomes more complicated, as the manipulator can contain other types of joints, such as prismatic and spherical joints and can contain passive revolute and prismatic joints. Kevin et al. \cite{cleary1994jacobian} formulated Jacobian for a novel 6-DOF parallel manipulator by splitting the mapping into two parts, one being the mapping between the end-effector forces/torques and the forces at the spherical joints, and the other being the mapping between the forces at the spherical joints and the torques at the actuating joints. Kim et al. \cite{846382} presented a formulation for obtaining analytic Jacobian of parallel manipulators using screw theory. Kim et al. \cite{kim2003new} presented the formulation of a new dimensionally homogeneous Jacobian matrix by taking three end-effector points. In their paper, the formulation of Jacobian is done by taking three points of the end-effector platform to interpolate any random point on the platform and differentiating it with respect to time, and finally rewriting the passive joint velocities in terms of active joint velocities and establishing the Jacobian. Geoffrey et al. \cite{pond2006formulating} also formulated a dimensionally homogeneous Jacobian matrix of parallel manipulators for dexterity analysis by using three points on the end-effector platform with a better physical significance of the properties of Jacobian. Liu et al. \cite{liu2010method} presented a method to systematically formulate the dimensionally homogeneous Jacobian when the manipulator has just one type of actuator, i.e., either prismatic or revolute. Their paper shows steps to formulate the generalised Jacobian by using the notation of screw theory. Hu \cite{hu2014formulation} presented a formulation of unified Jacobian for serial-parallel manipulators, i.e., a set of parallel manipulators linked one upon the other, serially. Even though there are many systematic formulations of Jacobian in the literature, there appears to be no unique method of systematic formulation of Jacobian that is applicable for both serial and parallel manipulators (including closed-loop spatial kinematic chains). In this study, a systematic formulation of Jacobian is presented that is applicable for both serial and parallel manipulators (including closed-loop spatial kinematic chains) containing four types of joints, namely revolute, prismatic, cylindrical and spherical.

Condition number of Jacobian is the ratio of maximum and minimum singular values of the Jacobian. It gives a measure of how much the manipulability is distributed to each singular value. Sometimes the manipulability measure alone might be unable to capture the information about singularity at a configuration in which case the condition number could give more information. For example, if a Jacobian has three singular values $100000$, $0.00002$ and $3$, the singular value $0.00002$ shows that it is close to singularity. But the manipulability would be $6$, from which it is not very clear that it is close to singularity, whilst the condition number would be $5\times 10^{9}$ which indicates that the manipulability is not equally distributed. Hence, condition numbers at the optimal configurations of the manipulators are important to analyse the performances. Jacobian is a mapping from joint velocities to end-effector velocities. This includes both the linear and the angular velocities of the end-effector that have different units. Similarly, the transpose of Jacobian maps the joint torques/forces to both forces and moments of the end-effector, which again have different units. This difference in units makes it difficult to assess the significance of the condition number of Jacobian. It is discussed in the literature that due to the non-uniform dimensions of the elements of the Jacobian matrix, the condition number may have little physical significance. Doty et al. \cite{388791} identified the problem of difference in units of elements of Jacobian and the difficulties associated with it. Angeles \cite{10.1177/027836499201100303} used the concept of natural length to present the Jacobian matrix in dimensionless form. In his paper, the natural length is chosen such that the condition number of the Jacobian matrix is minimised. Stocco et al. \cite{795800} used a scaling matrix with which the Jacobian matrix is to be multiplied in order to normalise the units of the Jacobian matrix and to use it with a performance goal. Their paper chooses the scaling matrix based on maximum desired forces. Ma et al. \cite{240404} discussed non-uniformity of units in the elements of Jacobian matrix, and used characteristic length to homogenise the Jacobian matrix. In their paper, a homogenised form of Jacobian matrix is presented for a 6-DOF platform manipulator, with characteristic length as a chosen parameter. In their paper, they suggested a scaling of the Jacobian matrix by multiplying it with the matrix $[S]=\begin{bmatrix}
\frac{1}{L} & 0 & 0 & 0 & 0 & 0\\
0 & \frac{1}{L} & 0 & 0 & 0 & 0\\
0 & 0 & \frac{1}{L} & 0 & 0 & 0\\
0 & 0 & 0 & 1 & 0 & 0\\
0 & 0 & 0 & 0 & 1 & 0\\
0 & 0 & 0 & 0 & 0 & 1\\
\end{bmatrix}$ where $L$ is called the characteristic length. As mentioned in their paper, there are many ways to define the characteristic length depending on the interest of the user. Various proposals of characteristic length were made in the literature, depending on the orientation of the problem that is to be solved. Their paper chose $L$, for platform manipulators of spherical joints, as the average of the distances from the centroid of the moving plate to each of the joints.

Yoshikawa \cite{yoshikawa1985manipulability} used the concept of singular values of Jacobian matrix as the manipulability measure to describe the performance of manipulators. Lee \cite{656551} used the concept of manipulability polytypes as a competing measure of manipulability. In his paper, the manipulability measure through manipulability ellipsoids is compared with that of manipulability polytopes. The paper concludes that the polytope approach can represent the manipulability more accurately than that of the ellipsoid approach, as the ellipsoid would not be covering the entire region of the set of unit joint velocities in the joint-space. Even though the polytope approach represents a better description of manipulability measure than the ellipsoid approach, the ellipsoid approach is simple to implement. To reduce the computational complexity and for simplicity, the ellipsoid approach is implemented in this study.

Khezrian et al. \cite{khezrian2014multi}, in their paper, designed a spherical 3-DoF parallel manipulator by maximising Global Dynamic Conditioning Index. In their paper, the Global Dynamic Conditioning Index of the manipulator is optimised, and the optimal dimensional parameters are presented. Hazarathaiah \cite{PappuriHazarathaiah} presented a prescription of several planar manipulators of revolute joints with corresponding optimised manipulability indices and condition numbers for a specified task position. In his thesis, several planar manipulators are presented, and the manipulability index of each manipulator is calculated as a function of its dimensional parameters. For each of the manipulators, the manipulability index is maximised, and the corresponding dimensions are presented. But this is limited to planar manipulators with revolute type of joints. The current study presents the prescriptions for several spatial manipulators with four types of joints, namely revolute, prismatic, cylindrical and spherical joints.

%

\section{Methodology}
\label{methodology}
\subsection{Methodology to formulate Jacobian}
\label{steps_to_formulate_jacobian}
The method used to generate Jacobian matrices for the manipulators in the context of this study is based on identifying all possible connecting paths from the base link to the end-effector link and kinematically formulating linear and angular velocities of the end-effector link through these paths, and finally transforming the passive joint velocities in terms of active joint velocities.

Identifying all possible connecting paths:

\begin{figure}[hbt!]
  \centering
  \includegraphics[width=\linewidth]{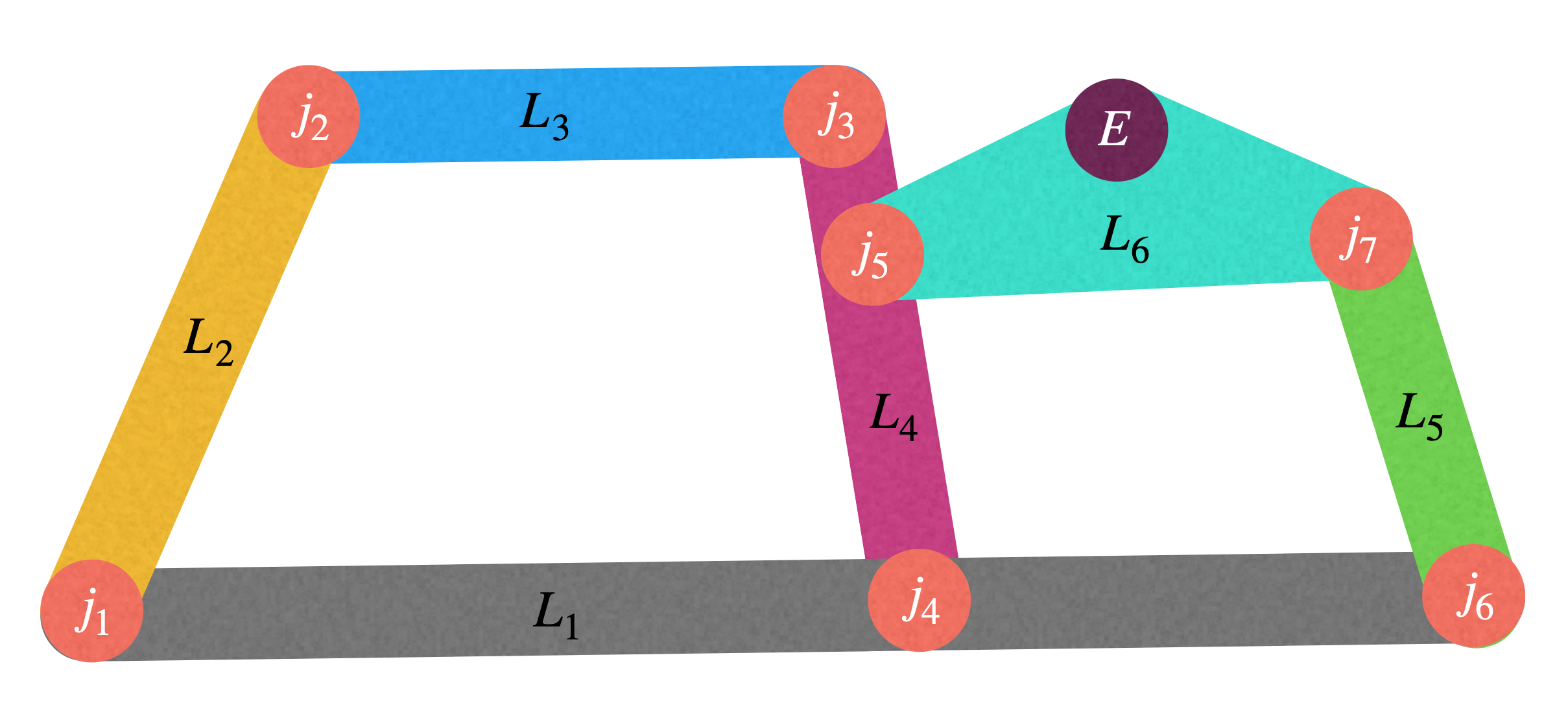}
  \caption{A mechanism with loops for illustration.}
  \label{fig_illustration_5}
\end{figure}

For the manipulator depicted in figure \ref{fig_illustration_5}, the connecting paths from the base link to the end-effector link are shown below.

\quad \quad Path 1: $L_1 - j_1 - L_2 - j_2 - L_3 - j_3 - L_4 - j_5 - L_6$

\quad \quad Path 2: $L_1 - j_4 - L_4 - j_5 - L_6$

\quad \quad Path 3: $L_1 - j_6 - L_5 - j_7 - L_6$

Formulating linear and angular velocities:

Each consecutive pair of links of the connecting path is considered to cumulatively formulate linear and angular velocities from the base link to the end-effector link, as each joint contributes linear and angular velocities to the end-effector link. For each consecutive pair of links in the connecting path, the contributions to the angular velocity and the linear velocity of the end-effector are given by $\vec{\omega}_{ij}$ and $\vec{v}_{ij}$ respectively in various cases, as shown in table \ref{Tab:relative_velocities}.


\begin{table}[h]  \centering \begin{tabular}{|c|c|c|}
\hline
Joint type & $\vec{\omega}_{ij}$ & $\vec{v}_{ij}$ \\ 
\hline
Revolute & $\dot{\theta}_{ij}\hat{n}_{ij}$ & $\dot{\theta}_{ij}\hat{n}_{ij}\times \left(\vec{a}-\vec{r}_{ij}\right)$ \\
Prismatic & 0 & $\dot{d}_{ij}\hat{n}_{ij}$ \\
Cylindrical & $\dot{\theta}_{ij}\hat{n}_{ij}$ & $\dot{\theta}_{ij}\hat{n}_{ij}\times \left(\vec{a}-\vec{r}_{ij}\right) + \dot{d}_{ij}\hat{n}_{ij}$ \\
Spherical & $\vec{\omega}_{ij}$ & $\vec{\omega}_{ij} \times \left( \vec{a} - \vec{r}_{ij}\right)$ \\
\hline\end{tabular} \caption{Result of the optimisation problem corresponding to 2D-M71 manipulator.} \label{Tab:relative_velocities} \end{table}





And the angular and linear velocities of the end-effector, described through the connecting path, are given by the sum of all the components of angular velocities and the sum of all the components of linear velocities contributed to the end-effector, respectively. Here, $\vec{r}_{ij}=\begin{Bmatrix} r_{ijx} & r_{ijy} & r_{ijz} \end{Bmatrix}^T$ and $\hat{n}_{ij}=\begin{Bmatrix} n_{ijx} & n_{ijy} & n_{ijz} \end{Bmatrix}^T$ represent the position vector of the instantaneous location and the axis of the instantaneous motion respectively, of the joint connected to the links $L_i$ \& $L_j$, and $\vec{\omega}_{ij}=\begin{Bmatrix} \omega_{ijx} & \omega_{ijy} & \omega_{ijz} \end{Bmatrix}^T$ represents the relative angular velocity vector of the link $L_j$ relative to the link $L_i$, of the adjacency matrix. $\vec{a}=\begin{Bmatrix} a_{x} & a_{y} & a_{z} \end{Bmatrix}^T$ is the position vector of the end-effector, and $\dot{\theta}_{ij}$ and $\dot{d}_{ij}$ represent the translational and the rotational joint velocities respectively, of the joint connected to the links $L_i$ and $L_j$ of the adjacency matrix, measured relative to the link $L_i$.

Applying the above process to each connecting path from base link to end-effector link gives the description of velocity of the end-effector in terms of linear combinations of active and passive joint velocities, the coefficients being functions of the configuration parameters of the manipulator, i.e., $r_{ijx}$, $r_{ijy}$, $r_{ijz}$, $\beta_{ij}$, $\phi_{ij}$, $a_x$, $a_y$ and $a_z$, where $\vec{a}=\begin{Bmatrix} a_x & a_y & a_z \end{Bmatrix}^T$ is the position vector of the end-effector point. For the particular kind of manipulator shown in figure \ref{fig_illustration_5}, there are three paths and hence three descriptions of pairs of angular and linear velocities can be formulated as $\begin{Bmatrix} v_1 \\ \omega_1 \end{Bmatrix}$, $\begin{Bmatrix} v_2 \\ \omega_2 \end{Bmatrix}$ and $\begin{Bmatrix} v_3 \\ \omega_3 \end{Bmatrix}$, each expressed in the form of linear combinations of active and passive velocities as shown in equation \eqref{eq:velocities_pair} for all $i$.

\begin{eqnarray}
    \label{eq:velocities_pair}
    \begin{Bmatrix} v^{(i)} \\ \omega^{(i)} \end{Bmatrix} = \left[J^{(i)}\right]\begin{Bmatrix} \dot{\theta}_a \\ \Omega \end{Bmatrix} = \left[J_1^{(i)}\right]\begin{Bmatrix}\dot{\theta}_a\end{Bmatrix}+\left[J_2^{(i)}\right]\begin{Bmatrix}\Omega\end{Bmatrix}
\end{eqnarray}

Each of $\begin{Bmatrix} v^{(i)} \\ \omega^{(i)} \end{Bmatrix}$ would represent the velocity components of the end-effector. Since the Jacobian matrix is a mapping of end-effector velocities with active joint velocities, all the passive joint velocities, i.e., all the variables $\dot{\theta}_{ij}$, $\dot{d}_{ij}$, $\omega_{ijx}$, $\omega_{ijy}$ and $\omega_{ijz}$ other than the actuating joint velocities, are to be written in terms of active joint velocities. The number of independent formulations of velocities (through the identified connecting paths from the base link to the end-effector link) should be more than or equal to the number of passive joint velocities that are to be eliminated. Once the passive joint velocities are written in terms of linear combinations of actuating joint velocities, the velocities of the end-effector can be expressed as a mapping of actuating joint velocities alone, and the mapping matrix thus generated would be the Jacobian matrix. The convention followed in this study in order to formulate the Jacobian matrix is to consider $\begin{Bmatrix} v_1 \\ \omega_1 \end{Bmatrix}$ as the pair of linear and angular velocities of the end-effector as shown in equation \eqref{eq:velocities_v1} and to consider $\begin{Bmatrix} v_i-v_1 \\ \omega_i-\omega_1 \end{Bmatrix}$ for all $i\neq 1$ to form the matrices $\left[A_1\right]$ and $\left[A_2\right]$, as shown in equation \eqref{eq:velocities_v2toN}, where $N$ is the number of paths from base link to end-effector link of the manipulator.

\begin{eqnarray}
    \label{eq:velocities_v1}
    \begin{Bmatrix} v \\ \omega \end{Bmatrix} = \begin{Bmatrix} v_1 \\ \omega_1 \end{Bmatrix} = \left[J_1\right]\begin{Bmatrix}\dot{\theta}_a\end{Bmatrix}+\left[J_2\right]\begin{Bmatrix}\Omega\end{Bmatrix}
\end{eqnarray}

\begin{eqnarray}
    \label{eq:velocities_v2toN}
    \begin{Bmatrix} v_2-v_1 \\ v_3-v_1 \\ \cdots \\ v_N-v_1 \\ \omega_2-\omega_1 \\ \omega_3-\omega_1 \\ \cdots \\ \omega_N-\omega_1 \end{Bmatrix} = \left[A\right]\begin{Bmatrix}\dot{\theta}\end{Bmatrix} = \left[A_1\right]\begin{Bmatrix}\dot{\theta}_a\end{Bmatrix}+\left[A_2\right]\begin{Bmatrix}\Omega\end{Bmatrix} = 0
\end{eqnarray}

From \eqref{eq:velocities_v2toN}, if $A_2$ is invertible, the passive joint velocities can be written in terms of active joint velocities as shown in \eqref{eq:thp_in_tha}.

\begin{eqnarray}
    \label{eq:thp_in_tha}
    \begin{Bmatrix}\Omega\end{Bmatrix} = -\left[A_2\right]^{-1}\left[A_1\right]\begin{Bmatrix}\dot{\theta}_a\end{Bmatrix}
\end{eqnarray}

By putting \eqref{eq:thp_in_tha} in \eqref{eq:velocities_v1}, the pair of linear and angular velocities of the end-effector can be written in terms of active joint velocities alone, as shown in \eqref{eq:velocity_in_tha}.


\begin{equation}
\begin{array}{l}
    \begin{Bmatrix} v \\ \omega \end{Bmatrix} = \left[J_1\right]\begin{Bmatrix}\dot{\theta}_a\end{Bmatrix}-\left[J_2\right]\left[A_2\right]^{-1}\left[A_1\right]\begin{Bmatrix}\dot{\theta}_a\end{Bmatrix} \\
    = \left(\left[J_1\right]-\left[J_2\right]\left[A_2\right]^{-1}\left[A_1\right]\right)\begin{Bmatrix}\dot{\theta}_a\end{Bmatrix} \\
    \label{eq:velocity_in_tha}
    = \left[\widetilde{J}\right]\begin{Bmatrix}\dot{\theta}_a\end{Bmatrix}
\end{array}
\end{equation}

From equation \eqref{eq:velocity_in_tha}, Jacobian of the manipulator can be defined as $\left[\widetilde{J}\right]=\left[J_1\right]-\left[J_2\right]\left[A_2\right]^{-1}\left[A_1\right]$ and the elements of Jacobian would be functions of configuration parameters, i.e., $r_{ijx}$, $r_{ijy}$, $r_{ijz}$, $n_{ijx}$, $n_{ijy}$, $n_{ijz}$, $a_x$, $a_y$ and $a_z$.

\subsubsection{The case of superfluous DOF}

\begin{figure}[hbt!]
  \centering
  \includegraphics[width=\linewidth]{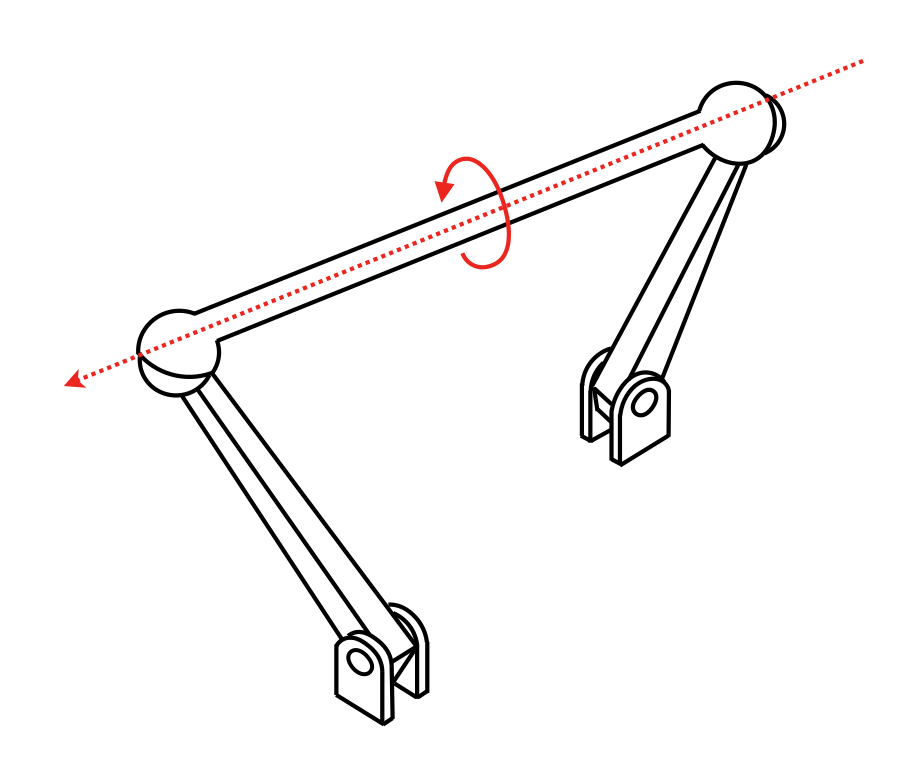}
  \caption{RSSR mechanism as an example of superfluous DOF.}
  \label{RSSR}
\end{figure}

If the system of equations $\left[A\right]\begin{Bmatrix}\dot{\theta}\end{Bmatrix}=0$ happens to have less equations than the number of unknowns then it signifies that the equations are inadequate to determine all the unknowns. Assuming that the right number of DOF is considered in the formulation, this could happen when the mechanism has a superfluous DOF. Superfluous DOF, in the context of this study, is the part of DOF of the mechanism that is not needed to impart the end-effector link to desired motion, which is the rotation of a link or a set of connected links of a mechanism, about an axis passing through the centres of two spherical joints, which does not alter either linear velocity or angular velocity of the end-effector. An example of superfluous DOF is the rotation of the link connected by two spherical joints about an axis that passes through the centres of the two spherical joints in the four-bar RSSR spatial manipulator, as shown in figure \ref{RSSR}.

An appropriate fix is made in this study that can enable the calculation of manipulability index. In the first step, the link (or the set of links) connected to the rest of the mechanism by two spherical joints alone, is identified. The angular velocity of the link (or the set of links) about the axis passing through the two spherical joints is to be set to zero. This can be achieved by the equation \eqref{eq:supfludofeqn}, where $\vec{\omega}_{k}$ is the absolute velocity of the link if it is the case of a single link (and is the absolute velocity of any link that is connected to one of the spherical joints if it is the case of a set of links), $\vec{r}_{ij}$ and $\vec{r}_{kl}$ are the position vectors of the two spherical joints.
\begin{eqnarray}
    \label{eq:supfludofeqn}
    \vec{\omega}_{k}\cdot \left(\vec{r}_{ij}-\vec{r}_{kl}\right) = 0
\end{eqnarray}

This fixes the issue and enables to calculate the manipulability index. Here, the superfluous angular velocity of the link is set to zero for simplicity, although setting it to any other value also fixes the issue.


\subsubsection{Type-2 singularity}
The formulation of Jacobian given by the equation \eqref{eq:velocity_in_tha} can be calculated only if $\left[A_2\right]$ is invertible. If it is not invertible, then the calculation of Jacobian would break down. Furthermore, any configuration close to such a configuration would end up having its Jacobian potentially badly scaled due to their singular values being close to zero. This is a type of singularity \cite{56660} widely known as Type-2 singularity \cite{10.1115/1.4001121,PAGIS20151,BRIOT2008445,7989721,7989720,6907477}, and this exists only in parallel manipulators (including closed-loop kinematic manipulators of single DOF, etc.), as $\left[A_2\right]$ does not come into picture in case of serial manipulators.

\subsection{Methodology to formulate condition number of a scaled matrix}
\label{methodology_of_scaled_matrix}

Since revolute, prismatic, cylindrical and spherical joints are used in the current study, the characteristic length used in this study is defined as follows.

If all the joints of a given manipulator are prismatic, then scaling is not needed.
Suppose the manipulator consists of revolute joints or cylindrical joints. In that case, an effective distance is considered for each joint, which is the shortest distance between the end-effector point and the axis of rotation of the revolute joint. If the position vector of the joint is $\vec{r}_{ij}$ and the end-effector point is $\vec{a}$ and the unit vector along the axis of the rotation is $\hat{n}_{ij}$ then the effective distance is given by $\bar{d}=\left|\left(\vec{r}_{ij}-\vec{a}\right)\times \hat{n}_{ij}\right|$.
If there are spherical joints in the manipulator, then in such case, the effective distance for each spherical joint is the distance between the end-effector point and the centre of the spherical joint. If the position vector of the joint is $\vec{r}_{ij}$ and the end-effector point is $\vec{a}$ then the effective distance is given by $\bar{d}=\left|\vec{r}_{ij}-\vec{a}\right|$. And finally, the characteristic length $L$ is calculated as the average of all these effective distances. Once the characteristic length is calculated, the scaling matrix $[S]$ can be formed and multiplied appropriately to the actual Jacobian matrix to get the scaled Jacobian matrix ($[J_S]=[S][J]$). And the condition number for this scaled Jacobian matrix $[J_S]$ is used for comparison of manipulators in this study.

The manipulators are sorted by their corresponding condition numbers in ascending order, and the corresponding indices are presented in the 2-DOF, 3-DOF and 4-DOF prescriptions by the name $i_{\kappa}$.

\subsection{Methodology to formulate optimisation problem}

In the current study, dimensional synthesis of all the enumerated manipulators is performed for the task position represented by the position vector of the end-effector point, which is arbitrarily chosen in this study as $\vec{a}=3\hat{i}+4\hat{j}+5\hat{k}$. This task position is used for all the manipulators here, for which dimensional synthesis is done. All the dimensions of manipulators in this study are in S.I. units.

The objective function is the manipulability index which is given by equation \ref{eq:1}.

The manipulators used in the current study have four types of joints: revolute, prismatic, cylindrical and spherical. When the applicable positions and orientations of the joints are defined, the dimensions of the manipulator can be determined. Thus, the objective function would be a function of the applicable positions and orientations of the joints.

The elements of Jacobian matrix would be functions of configuration parameters, i.e., $r_{ijx}$, $r_{ijy}$, $r_{ijz}$, $n_{ijx}$, $n_{ijy}$ and $n_{ijz}$. The manipulability measure of the manipulator is the product of singular values, which would be a scalar function of the configuration parameters. It can be calculated by the expression $\mu=\sqrt{\det{\left([J]^T[J]\right)}}$ if the number of actuating joints is less than or equal to the number of independent scalar quantities of the output velocity and $\mu=\sqrt{\det{\left([J][J]^T\right)}}$ otherwise. In the context of the current study, the spatial manipulators have up to six degrees of freedom. Since the enumerated list of manipulators in this study comprises 1-DOF, 2-DOF, 3-DOF, and 4-DOF manipulators, the number of actuating joints would be at most 4, which is less than 6. And hence $\mu=\sqrt{\det{\left([J]^T[J]\right)}}$ is considered for all the manipulators.

For simplicity in calculations, it is assumed that the locations of the joints should be within the cube of length $10 \text{m}$ in the first octave with one of its corners at origin. Therefore the optimisation problem for each of the manipulators can be defined as follows.

Maximise
\[\mu=\sqrt{\det{\left([J]^T[J]\right)}}\]
subject to constraints
\[n_{ijx}^2+n_{ijy}^2+n_{ijz}^2=1,\]
\[0\leq r_{ijx}\leq10,\]
\[0\leq r_{ijy}\leq10,\]
\[0\leq r_{ijz}\leq10.\]

In parallel manipulators, two new types of singularities arise, making the total number of types of singularities to be three. A normal mapping of end-effector velocities and joint velocities for parallel manipulators is typically expressed in the form as shown in equation \eqref{eq:type2expl1}.

\begin{eqnarray}
    \label{eq:type2expl1}
    \begin{Bmatrix}\dot{X}\end{Bmatrix}=\left(\left[J_1\right]-\left[J_2\right]\left[A_2\right]^{-1}\left[A_1\right]\right)\begin{Bmatrix}\dot{\theta}\end{Bmatrix}
\end{eqnarray}

If $\left[A_2\right]$ is singular then the product of singular values of Jacobian would become infinity. The physical significance of this is that at a particular set of dimensions the end-effector gains a DOF that cannot be controlled by the joint velocities. Equation \eqref{eq:type2expl1} can be written in a more simplified form as

\begin{eqnarray*}
    \det{\left(\left[A_2\right]\right)}\left[I\right]\begin{Bmatrix}\dot{X}\end{Bmatrix}+\left(\left[J_2\right]\text{adj}\left(\left[A_2\right]\right)\left[A_1\right]-\det{\left(\left[A_2\right]\right)}\left[J_1\right]\right)\begin{Bmatrix}\dot{\theta}\end{Bmatrix}=0 \\ 
    \Rightarrow \left[\widetilde{A}\right]\begin{Bmatrix}\dot{X}\end{Bmatrix}+\left[\widetilde{B}\right]\begin{Bmatrix}\dot{\theta}\end{Bmatrix}=0
\end{eqnarray*}

where $\left[\widetilde{A}\right]=\det{\left(\left[A_2\right]\right)}\left[I\right]$ and $\left[\widetilde{B}\right]=\left[J_2\right]\text{adj}\left(\left[A_2\right]\right)\left[A_1\right]-\det{\left(\left[A_2\right]\right)}\left[J_1\right]$.

The three types \cite{gosselin1990singularity} of singularities are outlined below.

\emph{Type 1 singularity:} When the product of singular values of $\left[\widetilde{B}\right]$ is zero, it is referred to as type 1 singularity, which is commonly encountered in serial manipulators and also parallel manipulators.

\emph{Type 2 singularity:} When the product of singular values of $\left[\widetilde{A}\right]$ is zero, it is referred to as type 2 singularity, which is encountered only in parallel manipulators.

\emph{Type 3 singularity:} When both the products of singular values of $\left[\widetilde{A}\right]$ and $\left[\widetilde{B}\right]$ are individually zero, it is referred to as type 3 singularity.

These are compactly summarised in table \ref{Tab:singularity_types}.

\begin{table}[h]  \centering 
{\renewcommand{\arraystretch}{2}
\begin{tabular}{|c|c|}
\hline
\bf{Singularity type} & \bf{Condition} \\ 
\hline
Type 1 & $\left[\widetilde{B}\right]$ is singular \\
\hline
Type 2 & $\left[\widetilde{A}\right]$ is singular \\
\hline
Type 3 & Both $\left[\widetilde{A}\right]$ and $\left[\widetilde{B}\right]$ are singular \\
\hline\end{tabular}}
\caption{Various types of singularities.} \label{Tab:singularity_types} \end{table}

Since a zero singular value of either of the matrices could lead to singularity, in order to design for optimal performance, the function $f_2=\sqrt{\det{\left(\left[\widetilde{A}\right]^T\left[\widetilde{A}\right]\right)}\det{\left(\left[\widetilde{B}\right]^T\left[\widetilde{B}\right]\right)}}$ is used \cite{PappuriHazarathaiah} in optimisation whenever $f_1=\sqrt{\det{\left(\left[\widetilde{J}\right]^T\left[\widetilde{J}\right]\right)}}$ of the manipulator becomes infinity during optimisation process.

\subsection{Strategy used in the optimisation process}
\label{strategy_used_in_optimisation}
Since classical optimisation algorithms can find only local minima, multiple initial guesses are considered to reasonably attempt to find global optima. The following strategy is used with consideration of identifying and handling the cases involving type 2 and type 3 singularities.

For each manipulator, the following four steps are performed.

\quad \emph{Step 1:} Maximisation of $f_1=\sqrt{\det{\left(\left[J\right]^T\left[J\right]\right)}}$ with 100 random initial guesses within the bounds. The local optimal points for which the condition number of $\left[\widetilde{A}\right]$ is less than 1000 and the minimum singular value of $\left[\widetilde{A}\right]$ is greater than $10^{-2}$, are shortlisted. For serial manipulators $\left[\widetilde{A}\right]$ does not exist and hence this step is omitted for serial manipulators.

\quad \emph{Step 2:} The maximum of these shortlisted points is considered to be the optimal point.

\quad \emph{Step 3:} If no point passes the shortlisting criteria, then it is assumed that type 2 or type 3 singularity has occurred. And hence, maximisation of $f_2=\sqrt{\det{\left(\left[\widetilde{A}\right]^T\left[\widetilde{A}\right]\right)\det{\left(\left[\widetilde{B}\right]^T\left[\widetilde{B}\right]\right)}}}$ is performed with a random initial guess.

\quad \emph{Step 4:} In step 3, for the local optimum point, if the condition number of Jacobian is less than 1000 and the minimum singular value of Jacobian is greater than $10^{-2}$, then the point is considered as the optimal point. If the condition number is greater than or equal to 1000 or if the minimum singular value is less than or equal to $10^{-2}$ or if the initial guess does not converge, then this initial guess is discarded, and a new random initial guess is used to repeat step 3 until the initial guess converges, the condition number ends up to be less than 1000, and the minimum singular value ends up to be greater than $10^{-2}$.

The built-in command \emph{fmincon} of MATLAB Optimisation Toolbox \cite{MatlabOTB}, with default options, is used to perform optimisation in this study. The default method that the command uses for this kind of optimisation problems is \emph{interior-point} with BFGS approximation of Hessian. The interior-point method is used to handle constrained optimisation, which uses a barrier function that pressurises the point during the iterations to lie within the boundary of the feasible space.

Some parameters used by the MATLAB function are

\quad Constraint Tolerance: $10^{-6}$


\quad Finite Difference Type: Forward

\quad Function Tolerance: $10^{-6}$

\quad Step Tolerance: $10^{-10}$


\section{Dimensional synthesis}
From a previous companion study \cite{https://doi.org/10.48550/arxiv.2210.03327} on the enumeration of manipulators, 96 1-DOF manipulators amongst 4 links and 645 2-DOF, 8 3-DOF and 15 4-DOF manipulators amongst 3, 4 and 5 links are considered for dimensional synthesis. All joints are actuated in case of serial manipulators, whereas there exist passive joints in case of closed-loop manipulators. Appropriate actuating joints are arbitrarily chosen from revolute and prismatic joints of the closed-loop manipulators. Dimensional synthesis for all the manipulators is performed based on the methodology presented in section \ref{methodology}.

As an example, detailed steps for dimensional synthesis of the manipulator 2D-M71 is shown below.

\subsection{Detailed formulation of the manipulator 2D-M71 as an example}

\input{contentfolder/mechanisms2/manipulators/dof2/manipulator_71_edited}

Similarly, manipulability indices and condition numbers (wherever applicable) for all the other manipulators are computed, and finally the prescriptions for 1-DOF, 2-DOF, 3-DOF and 4-DOF manipulators for optimal performance around the task point $\vec{a}=\{1,2,3\}^T$ are presented below.




\subsection{Prescriptions}

A list of manipulators with their manipulability indices $\left(\mu\right)$ and the corresponding scaled manipulability indices $\left(\bar{\mu}\right)$, is presented for each of 1-DOF, 2-DOF, 3-DOF and 4-DOF manipulators, in which the manipulators are sorted by scaled manipulability index in the descending order. For manipulators with more than 1-DOF, the corresponding condition numbers $\left(\kappa\right)$ and the corresponding indices $i_{\kappa}$ are also shown. The corresponding objective function used in optimisation is also shown with the name $f$. The designation $f=f_1$ implies that the objective function used is $f_1=\sqrt{\det{\left(\left[\widetilde{J}\right]^T\left[\widetilde{J}\right]\right)}}$ and the designation $f=f_2$ implies that the objective function used is $f_2=\sqrt{\det{\left(\left[\widetilde{A}\right]^T\left[\widetilde{A}\right]\right)}\times \det{\left(\left[\widetilde{B}\right]^T\left[\widetilde{B}\right]\right)}}$.

\subsubsection{Prescription for 1-DOF manipulators}



The prescription containing 96 1-DOF manipulators is shown in table 2. The last column, namely AJV (Active Joint Velocities), shows the joints that are considered to be actuating joints for the closed-loop manipulators. Since all the manipulators considered in this section are single-degree-of-freedom manipulators, the condition number of every manipulator would be 1 and hence the condition numbers are not shown in this prescription.

\input{contentfolder/mechanisms2/1dof_manipulabilitytable_final}

This prescription gives the designer the set of 1-DOF manipulators arranged in descending order of performance of manipulators to operate around the end-effector point $\vec{a}=\{3,4,5\}^T$.

\subsubsection{Prescription for 2-DOF manipulators}

A contracted version of the prescription containing 645 1-DOF manipulators is shown in table 3. Full table is shown in appendix A, in which the actuating joint velocities (AJV) considered for this study are shown. In the table, $\kappa$ refers to the condition number of the Jacobian at the optimum point and $i_{\kappa}$ refers to the ranked position of the manipulator based on its condition number. 

\begin{center}
\tablefirsthead{%
\hline
\multicolumn{1}{|c|}{S.No.} &
Name &
$\bar{\mu}$ &
$\kappa$ &
$i_{\kappa}$ &
$\mu$ &
\multicolumn{1}{c|}{$f$} \\
\hline}
\tablehead{%
\hline
\multicolumn{7}{|l|}{\small\sl continuing...}\\
\hline
\multicolumn{1}{|c|}{S.No.} &
Name &
$\bar{\mu}$ &
$\kappa$ &
$i_{\kappa}$ &
$\mu$ &
\multicolumn{1}{c|}{$f$} \\
\hline}
\tabletail{%
\hline
\multicolumn{7}{|r|}{\small\sl ...continued}\\
\hline}
\tablelasttail{\hline}
\bottomcaption{Prescription of 2-DOF Manipulators.}
\begin{supertabular}{|r|r|r|r|r|r|r|}
\hline
1 & 2D-M645 & 1.0 & 1.0 & 6 & 1.0 & $f_1$ \\
\hline2 & 2D-M315 & 0.7544 & 1.2 & 155 & 87.4 & $f_1$ \\
\hline3 & 2D-M301 & 0.7058 & 1.2 & 158 & 88.8 & $f_2$ \\
\hline4 & 2D-M224 & 0.6366 & 1.1 & 127 & 60.2 & $f_1$ \\
\hline5 & 2D-M308 & 0.6034 & 1.1 & 118 & 340.6 & $f_2$ \\
\hline6 & 2D-M126 & 0.5855 & 1.3 & 207 & 88.8 & $f_1$ \\
\hline7 & 2D-M309 & 0.5624 & 1.2 & 135 & 351.6 & $f_2$ \\
\hline8 & 2D-M319 & 0.5621 & 1.4 & 230 & 283.4 & $f_2$ \\
\hline9 & 2D-M208 & 0.5495 & 1.2 & 172 & 68.1 & $f_1$ \\
\hline10 & 2D-M134 & 0.541 & 1.1 & 107 & 100.5 & $f_1$ \\
\hline11 & 2D-M248 & 0.5381 & 1.4 & 224 & 75.2 & $f_1$ \\
\hline12 & 2D-M203 & 0.5289 & 1.3 & 198 & 68.1 & $f_1$ \\
\hline13 & 2D-M634 & 0.5142 & 1.0 & 7 & 1 & $f_1$ \\
\hline14 & 2D-M458 & 0.5116 & 1.0 & 32 & 1.0 & $f_1$ \\
\hline15 & 2D-M74 & 0.5114 & 1.1 & 115 & 77.5 & $f_1$ \\
\hline\mathcolorbox{\vdots} & \mathcolorbox{\vdots} & \mathcolorbox{\vdots} & \mathcolorbox{\vdots} & \mathcolorbox{\vdots} & \mathcolorbox{\vdots} & \mathcolorbox{\vdots} \\
\hline641 & 2D-M579 & 0.0007 & 1.4 & 243 & 0.5 & $f_2$ \\
\hline642 & 2D-M591 & 0.0007 & 1.4 & 256 & 0.5 & $f_2$ \\
\hline643 & 2D-M572 & 0.0007 & 1.4 & 239 & 0.5 & $f_2$ \\
\hline644 & 2D-M491 & 0.0007 & 1.3 & 203 & 0.5 & $f_2$ \\
\hline645 & 2D-M607 & 0.0006 & 1.6 & 282 & 0.4 & $f_2$ \\
\hline

\end{supertabular}
\end{center}

This prescription gives the designer the set of 2-DOF manipulators arranged in descending order of performance of manipulators to operate around the end-effector point $\vec{a}=\{3,4,5\}^T$.

\subsubsection{Prescription for 3-DOF manipulators}


The prescription containing 8 3-DOF manipulators is shown in table \ref{Tab:table_dof3mt_1}. All the eight manipulators happened to be serial manipulators and hence the objective function $f$ is $f_1$ for all the 8 manipulators. Since all the 3-DOF manipulators here are serial manipulators, all the joints would be actuating joints, and hence the AJV column is not shown in the table.

\input{contentfolder/mechanisms2/manipulabilitytable_new3}

This prescription gives the designer the set of 3-DOF manipulators arranged in descending order of performance of manipulators to operate around the end-effector point $\vec{a}=\{3,4,5\}^T$.

\subsubsection{Prescription for 4-DOF manipulators}

The prescription containing 15 4-DOF manipulators is shown in table \ref{Tab:table_dof3mt_1}. All the 15 manipulators also happened to be serial manipulators and hence the objective function $f$ is $f_1$ for all the manipulators. Since all these manipulators are serial manipulators, all the joints would be actuating joints, and hence the AJV column is not shown in the table.

\input{contentfolder/mechanisms2/manipulabilitytable_new4}

This prescription gives the designer the set of 4-DOF manipulators arranged in descending order of performance of manipulators to operate around the end-effector point $\vec{a}=\{3,4,5\}^T$. 

\section{Discussions}

\subsection{On the prescription for manipulators of DOF 1}
The manipulator with the highest scaled manipulability is $\text{1D-M10}$ with $\bar{\mu}=0.1523$ and the corresponding dimensional parameters from the corresponding optimisation result are


$$\begin{matrix}
\hat{n}_{14}=-\hat{k} \\
\hat{n}_{23}=0.04\hat{i}-0.71\hat{j}+0.7\hat{k} \\
\hat{n}_{24}=-0.84\hat{i}-0.49\hat{j}-0.25\hat{k} \\
\vec{r}_{13}=3.76\hat{i}+5.51\hat{j}+5.34\hat{k} \\
\vec{r}_{14}=10\hat{i}+1.6\hat{k} \\
\vec{r}_{23}=4.11\hat{i}+6.4\hat{j}+4.81\hat{k} \\
\vec{r}_{24}=4.67\hat{i}+3.62\hat{j}+5.9\hat{k}
\end{matrix}
$$

From the above data, the link lengths can be derived as follows.

$$\begin{matrix}
l_{1}=|\vec{r}_{13}-\vec{r}_{14}|=9.13 \\
l_{2}=|\vec{r}_{23}-\vec{r}_{24}|=3.04 \\
l_{3}=|\vec{r}_{13}-\vec{r}_{23}|=1.09 \\
l_{4a}=|\vec{r}_{14}-\vec{r}_{24}|=7.75 \\
l_{4b}=|\vec{r}_{14}-\vec{a}|=8.75 \\
l_{4c}=|\vec{r}_{24}-\vec{a}|=1.94
\end{matrix}
$$

Hence, the first link should have a length of $9.13$m, the second link should have a length of $3.04$m, the third link should have a length of $1.09$m, and the forth link, if assumed to be of triangular shape, should have its sides $7.75$m, $8.75$m and $1.94$m, wherein the sides of lengths $8.75$m and $1.94$m meet at the end-effector point, and the side of length $8.75$m connects the end-effector and the revolute joint.

The revolute joint connecting the links $2$ and $3$ is to be located at $(4.11, 6.4, 4.81)$ with its axis oriented along the unit vector $0.04\hat{i}-0.71\hat{j}+0.7\hat{k}$. The other revolute joint is to be located at $(10, 0, 1.6)$ with its axis oriented along the unit vector $-\hat{k}$. The cylindrical joint is to be located at $(4.67,3.62,5.9)$ with its axis oriented along the unit vector $-0.84\hat{i}-0.49\hat{j}-0.25\hat{k}$. Finally, the spherical joint is to be located at $(3.76,5.51,5.34)$.

If the designer, for some other constraints, is not interested in this particular type of manipulator, i.e., $\text{1D-M10}$, the prescription provides the next best manipulator, i.e., $\text{1D-M65}$, for which the corresponding dimensional parameters are available. This gives the designer the set of 1-DOF manipulators to choose the manipulator from, with the appropriate dimensions that are necessary to build the manipulator.

\subsection{On the prescription for manipulators of DOF 2}
The manipulator with the highest scaled manipulability is $\text{2D-M645}$ with $\bar{\mu}=1.0$ and the corresponding dimensional parameters are

$$\begin{matrix}
    \hat{n}_{12}=0.97\hat{i}+0.03\hat{j}+0.24\hat{k} \\
    \hat{n}_{23}=0.03\hat{i}-1.0\hat{j}-0.01\hat{k} \\
\end{matrix}
$$

This shows that the axis of the prismatic joint connecting the links $1$ and $2$ is to be along the unit vector $0.97\hat{i}+0.03\hat{j}+0.24\hat{k}$ and the axis of the prismatic joint connecting the links $2$ and $3$ is to be along the unit vector $0.03\hat{i}-1.0\hat{j}-0.01\hat{k}$.

In the optimised result, there are no specified locations for the prismatic joints. This is because the Jacobian is not dependent on the locations of prismatic joints, as a prismatic joint can give the same velocity regardless of the location of its joint, as long as its axis is unchanged. Hence, the parameters corresponding to locations of the prismatic joints, i.e., the components of the position vectors $\vec{r}_{12}$ and $\vec{r}_{23}$, are free parameters that can be chosen by the designer within the specified bounds.


For any reason if the designer does not opt for this particular type of manipulator, i.e., $\text{2D-M645}$, the next best manipulator is provided in the prescription, which is $\text{2D-M315}$, along with the corresponding dimensional parameters from the optimisation results. This gives the designer the list of 2-DOF manipulators to choose the manipulator from, with the appropriate dimensional parameters that are essential to build the manipulator.
\subsection{On the prescription for manipulators of DOF 3}
The manipulator with the highest scaled manipulability is $\text{3D-M1}$ with $\bar{\mu}=1.0113$ and the corresponding dimensional parameters are

$$\begin{matrix}

    \hat{n}_{13}=-0.71\hat{i}+0.11\hat{j}-0.7\hat{k} \\ 
    \hat{n}_{23}=-0.0\hat{i}+0.58\hat{j}-0.81\hat{k} \\ 
    \hat{n}_{24}=0.82\hat{i}-0.46\hat{j}-0.34\hat{k} \\
    \vec{r}_{13}=10.0\hat{i}+10.0\hat{j}+0.0\hat{k} \\
    \vec{r}_{23}=10.0\hat{i}+0.0\hat{j}+10.0\hat{k} \\
    \vec{r}_{24}=10.0\hat{i}+10.0\hat{j}+10.0\hat{k}
    \end{matrix}
    $$

    From the above data, the link lengths can be derived as follows.

$$\begin{matrix}
l_{2}=|\vec{r}_{23}-\vec{r}_{24}|=14.14 \\
l_{3}=|\vec{r}_{13}-\vec{r}_{23}|=10.0 \\
l_{4}=|\vec{a}-\vec{r}_{24}|=10.49
\end{matrix}
$$

Hence, the second link should have a length of $14.14$m, the third link should have a length of $10.0$m, and the third link should have a length of $10.49$m.
    
    The revolute joint connecting the links $1$ and $3$ is to be located at $(10.0, 10.0, 0.0)$ with its axis oriented along the unit vector $-0.71\hat{i}+0.11\hat{j}-0.7\hat{k}$. The revolute joint connecting the links $2$ and $3$ is to be located at $(10.0, 0.0, 10.0)$ with its axis oriented along the unit vector $0.0\hat{i}+0.58\hat{j}-0.81\hat{k}$. And the revolute joint connecting the links $2$ and $4$ is to be located at $(10.0, 10.0, 10.0)$ with its axis oriented along the unit vector $0.82\hat{i}-0.46\hat{j}-0.34\hat{k}$.

    
    If the designer due to some other restrictions does not find the manipulator $\text{3D-M1}$ viable, the next best manipulator, i.e., $\text{3D-M8}$, is available in the prescription, with its corresponding dimensional parameters. This gives the designer an atlas of 3-DOF manipulators to choose the manipulator from, with the appropriate dimensional parameters that are necessary to build the manipulator.

\subsection{On the prescription for manipulators of DOF 4}
The manipulator with the highest scaled manipulability is $\text{4D-M1}$ with $\bar{\mu}=2.5607$ and the corresponding dimensional parameters are

$$\begin{matrix}
    \hat{n}_{14}=-0.29\hat{i}+0.57\hat{j}-0.77\hat{k} \\ 
    \hat{n}_{23}=0.75\hat{i}-0.53\hat{j}+0.4\hat{k} \\
    \hat{n}_{25}=-0.34\hat{i}+0.83\hat{j}-0.45\hat{k} \\
    \hat{n}_{34}=-0.72\hat{i}+0.12\hat{j}+0.69\hat{k} \\
    \vec{r}_{14}=10.0\hat{i}+0.0\hat{j}+10.0\hat{k} \\
    \vec{r}_{23}=10.0\hat{i}+10.0\hat{j}+0.0\hat{k} \\
    \vec{r}_{25}=0.0\hat{i}+0.0\hat{j}+0.0\hat{k} \\ 
    \vec{r}_{34}=10.0\hat{i}+10.0\hat{j}+10.0\hat{k}
\end{matrix}
$$

From the above data, the link lengths can be derived as follows.

$$\begin{matrix}
l_{4}=|\vec{r}_{14}-\vec{r}_{34}|=10.0 \\
l_{3}=|\vec{r}_{34}-\vec{r}_{23}|=10.0 \\
l_{2}=|\vec{r}_{23}-\vec{r}_{25}|=14.14 \\
l_{5}=|\vec{r}_{25}-\vec{a}|= 7.07\\
\end{matrix}
$$

Hence, the second link should have a length of $14.14$m, the third link should have a length of $10.0$m, the fourth link should have a length of $10.0$m, and the fifth link should have a length of $7.07$m.

The revolute joint connecting the links $1$ and $4$ is to be located at $(10.0, 0.0, 10.0)$ with its axis oriented along the unit vector $-0.29\hat{i}+0.57\hat{j}-0.77\hat{k}$. The revolute joint connecting the links $3$ and $4$ is to be located at $(10.0, 10.0, 10.0)$ with its axis oriented along the unit vector $-0.72\hat{i}+0.12\hat{j}+0.69\hat{k}$. The revolute joint connecting the links $2$ and $3$ is to be located at $(10.0, 10.0, 0.0)$ with its axis oriented along the unit vector $0.75\hat{i}-0.53\hat{j}+0.4\hat{k}$. And the revolute joint connecting the links $2$ and $5$ is to be located at $(0.0, 0.0, 0.0)$ with its axis oriented along the unit vector $-0.34\hat{i}+0.83\hat{j}-0.45\hat{k}$.

In case the designer could not use $\text{4D-M1}$ for any other constraint, the prescription provides the next best manipulator, i.e., $\text{4D-M13}$, with the corresponding dimensional parameters that are available from the optimisation results. This gives the designer the set of 4-DOF manipulators to choose the manipulator from, with the appropriate dimensions that are required to build the manipulator.

\section{Conclusion}

Dimensional synthesis is done for 1-DOF, 2-DOF, 3-DOF and 4-DOF manipulators, by optimising the manipulability index around the chosen end-effector point for this study. Condition numbers for 2-DOF, 3-DOF and 4-DOF manipulators are calculated. The DOF-wise performance of the manipulators is compared by using the scaled manipulability indices ($\bar{\mu}$) of the manipulators, and DOF-wise prescriptions are provided for manipulators. For 1-DOF, 1D-M10 is found to have the maximum scaled manipulability index. Among 2-DOF manipulators, 2D-M645 manipulator is found to have the maximum scaled manipulability index and 78 manipulators are found to be having the lowest condition number (rounded up to one decimal point). Among 3-DOF manipulators, 3D-M1 is found to have the maximum scaled manipulability index and 3D-M8 is found to have the lowest condition number. Among 4-DOF manipulators, 4D-M1 is found to have both the highest scaled manipulability index and the lowest condition number. 

The manipulator with the most scaled manipulability reflects the best performance around the required task position, provided the condition number is reasonable. In case the designer, due to some other constraints, is not interested in the manipulator with the highest scaled manipulability, the prescription provides the second best manipulator (and, then others ranked appropriately) with its corresponding condition number, if applicable. This, along with the result of the companion study, gives the designer an atlas of manipulators to choose the manipulator from, with the appropriate dimensions that are necessary to build the manipulator to operate around the specified task point.

\bibliography{sample}
\input{Appendices/Appendix_main}

\end{document}

%% file: contentfolder/mechanisms2/manipulators/dof2/manipulator_71_edited.tex

\begin{figure}[hbt!]
  \centering
  \includegraphics[width=0.6\linewidth]{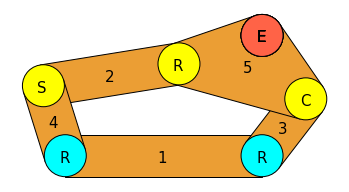}
  \caption{Schematic figure of 2D-M71.}
  \label{fig_2dof_iext_71}
\end{figure}

The schematic diagram of 2D-M71 manipulator is shown in figure \ref{fig_2dof_iext_71}, mentioning the actuator joints in cyan circles.

\noindent\emph{Formulation of Jacobian:}

Jacobian of the manipulator is formulated using the steps shown in section \ref{steps_to_formulate_jacobian}. The various paths that connect the base link and the end-effector link are $L_1-R-L_3-C-L_5$ and $L_1-R-L_4-S-L_2-R-L_5$. The descriptions of linear and angular velocities of the end-effector is formulated using various paths from the end-effector link to the base link, and shown by the equations below.


\begin{equation}
\begin{array}{cc}
    \vec{v}=\dot{\theta}_{13} \hat{n}_{13} \times \left( \vec{a} - \vec{r}_{13} \right) + \dot{\theta}_{35} \hat{n}_{35} \times \left( \vec{a} - \vec{r}_{35} \right) \\
     + \dot{d}_{35} \hat{n}_{35}
\end{array}
\label{eqn:eqn_2_71_1_2}
\end{equation}

\begin{equation}
\begin{array}{cc}
    \vec{v}=\dot{\theta}_{14} \hat{n}_{14} \times \left( \vec{a} - \vec{r}_{14} \right) + \vec{\omega}_{42} \times \left( \vec{a} - \vec{r}_{42} \right) \\
    + \dot{\theta}_{25} \hat{n}_{25} \times \left( \vec{a} - \vec{r}_{25} \right)
\end{array}
\label{eqn:eqn_2_71_1_1}
\end{equation}

\begin{dmath}\label{eqn:eqn_2_71_2_2}\vec{\omega}=\dot{\theta}_{13} \hat{n}_{13} + \dot{\theta}_{35} \hat{n}_{35}\end{dmath}\begin{dmath}\label{eqn:eqn_2_71_2_1}\vec{\omega}=\dot{\theta}_{14} \hat{n}_{14} + \vec{\omega}_{42} + \dot{\theta}_{25} \hat{n}_{25}\end{dmath}where $\vec{r}_{ij}=\begin{Bmatrix} r_{ijx} \\ r_{ijy} \\ r_{ijz} \end{Bmatrix}$ and $\hat{n}_{ij}=\begin{Bmatrix} n_{ijx} \\ n_{ijy} \\ n_{ijz} \end{Bmatrix}$ represent the position vector of the instantaneous location and the axis of the instantaneous motion respectively, of the joint connected to the links $L_i$ \& $L_j$, and $\vec{\omega}_{ij}=\begin{Bmatrix} \omega_{ijx} \\ \omega_{ijy} \\ \omega_{ijz} \end{Bmatrix}$ represents the relative angular velocity vector of the link $L_j$ relative to the link $L_i$. $\vec{a}=\begin{Bmatrix} a_{x} \\ a_{y} \\ a_{z} \end{Bmatrix}$ is the position vector of the end-effector, and $\dot{\theta}_{ij}$ and $\dot{d}_{ij}$ represent the translational and the rotational joint velocities respectively, of the joint connected to the links $L_i$ and $L_j$, measured relative to the link $L_i$.
\newline
From equations \eqref{eqn:eqn_2_71_1_2} and \eqref{eqn:eqn_2_71_2_2}, the vector of end-effector velocities in terms of both active and passive joint angles is given by
\newline
\resizebox{0.4\textwidth}{!}{$
$$\begin{Bmatrix}\vec{v} \\ \vec{\omega} \end{Bmatrix}=\left[\begin{matrix}n_{13y} \left(a_z - r_{13z}\right) - n_{13z} \left(a_y - r_{13y}\right) & 0 & n_{35x} & 0 \\- n_{13x} \left(a_z - r_{13z}\right) + n_{13z} \left(a_x - r_{13x}\right) & 0 & n_{35y} & 0 \\n_{13x} \left(a_y - r_{13y}\right) - n_{13y} \left(a_x - r_{13x}\right) & 0 & n_{35z} & 0 \\n_{13x} & 0 & 0 & 0 \\n_{13y} & 0 & 0 & 0 \\n_{13z} & 0 & 0 & 0 \end{matrix}\right.
$}
\\\qquad
\resizebox{0.4\textwidth}{!}{$
\left.\begin{matrix} n_{35y} \left(a_z - r_{35z}\right) - n_{35z} \left(a_y - r_{35y}\right) & 0 & 0 & 0\\ - n_{35x} \left(a_z - r_{35z}\right) + n_{35z} \left(a_x - r_{35x}\right) & 0 & 0 & 0\\ n_{35x} \left(a_y - r_{35y}\right) - n_{35y} \left(a_x - r_{35x}\right) & 0 & 0 & 0\\ n_{35x} & 0 & 0 & 0\\ n_{35y} & 0 & 0 & 0\\ n_{35z} & 0 & 0 & 0\end{matrix}\right]\begin{Bmatrix}\dot{\theta}_{13}\\\dot{\theta}_{14}\\\dot{d}_{35}\\\dot{\theta}_{25}\\\dot{\theta}_{35}\\\omega_{42x}\\\omega_{42y}\\\omega_{42z}\end{Bmatrix}$$
$}

\resizebox{0.45\textwidth}{!}{$
$$\Rightarrow \begin{Bmatrix}\vec{v} \\ \vec{\omega} \end{Bmatrix}=\left[\begin{matrix}n_{13y} \left(a_z - r_{13z}\right) - n_{13z} \left(a_y - r_{13y}\right) & 0\\- n_{13x} \left(a_z - r_{13z}\right) + n_{13z} \left(a_x - r_{13x}\right) & 0\\n_{13x} \left(a_y - r_{13y}\right) - n_{13y} \left(a_x - r_{13x}\right) & 0\\n_{13x} & 0\\n_{13y} & 0\\n_{13z} & 0\end{matrix}\right]\begin{Bmatrix}\dot{\theta}_{13}\\\dot{\theta}_{14}\end{Bmatrix}+$$
$}
\\\qquad
\resizebox{0.45\textwidth}{!}{$
$$\left[\begin{matrix}n_{35x} & 0 & n_{35y} \left(a_z - r_{35z}\right) - n_{35z} \left(a_y - r_{35y}\right) & 0 & 0 & 0\\n_{35y} & 0 & - n_{35x} \left(a_z - r_{35z}\right) + n_{35z} \left(a_x - r_{35x}\right) & 0 & 0 & 0\\n_{35z} & 0 & n_{35x} \left(a_y - r_{35y}\right) - n_{35y} \left(a_x - r_{35x}\right) & 0 & 0 & 0\\0 & 0 & n_{35x} & 0 & 0 & 0\\0 & 0 & n_{35y} & 0 & 0 & 0\\0 & 0 & n_{35z} & 0 & 0 & 0\end{matrix}\right]\begin{Bmatrix}\dot{d}_{35}\\\dot{\theta}_{25}\\\dot{\theta}_{35}\\\omega_{42x}\\\omega_{42y}\\\omega_{42z}\end{Bmatrix}$$
$}

\[\Rightarrow \begin{Bmatrix}\vec{v} \\ \vec{\omega} \end{Bmatrix}=\left[J_1\right]\{\dot{\theta}_a\}+\left[J_2\right]\{\Omega\}\]where $\{\dot{\theta}_a\}=\begin{Bmatrix}\dot{\theta}_{13}\\\dot{\theta}_{14}\end{Bmatrix}$ and $\{\Omega\}=\begin{Bmatrix}\dot{d}_{35} \\ \dot{\theta}_{25} \\ \dot{\theta}_{35} \\ \omega_{42x} \\ \omega_{42y} \\ \omega_{42z}\end{Bmatrix}$ represent the active joint-velocities and the passive joint-velocities, respectively.
\newline
But since the Jacobian should contain the end-effector velocities in terms of active joint angles alone, the passive joint angles must be written in terms of active joint angles. From the two equations \eqref{eqn:eqn_2_71_1_1} and \eqref{eqn:eqn_2_71_1_2},
\newline

\begin{equation}
\begin{array}{cc}
    \dot{\theta}_{13} \hat{n}_{13} \times \left( \vec{a} - \vec{r}_{13} \right) + \dot{\theta}_{35} \hat{n}_{35} \times \left( \vec{a} - \vec{r}_{35} \right) \\
    + \dot{d}_{35} \hat{n}_{35}=\dot{\theta}_{14} \hat{n}_{14} \times \left( \vec{a} - \vec{r}_{14} \right) \\
    + \vec{\omega}_{42} \times \left( \vec{a} - \vec{r}_{42} \right) + \dot{\theta}_{25} \hat{n}_{25} \times \left( \vec{a} - \vec{r}_{25} \right)
\end{array}
\label{eqn:eqn_2_71_5_1}
\end{equation}

From the two equations \eqref{eqn:eqn_2_71_2_1} and \eqref{eqn:eqn_2_71_2_2},
\newline
\begin{dmath}
  \label{eqn:eqn_2_71_5_2}
  \dot{\theta}_{14} \hat{n}_{14} + \vec{\omega}_{42} + \dot{\theta}_{25} \hat{n}_{25}=\dot{\theta}_{13} \hat{n}_{13} + \dot{\theta}_{35} \hat{n}_{35}\end{dmath}
From the equations \eqref{eqn:eqn_2_71_5_1} and \eqref{eqn:eqn_2_71_5_2}, linear equations in terms of active and passive joint angles can be stacked as shown below, using which the passive joint angles can be written in terms of the active joint angles.\newline

  \resizebox{0.3\textwidth}{!}{$
    $$\left[\begin{matrix}- n_{13y} \left(a_z - r_{13z}\right) + n_{13z} \left(a_y - r_{13y}\right) \\n_{13x} \left(a_z - r_{13z}\right) - n_{13z} \left(a_x - r_{13x}\right) \\- n_{13x} \left(a_y - r_{13y}\right) + n_{13y} \left(a_x - r_{13x}\right) \\- n_{13x} \\- n_{13y} \\- n_{13z} \end{matrix}\right.$$
    $}
    \\\qquad
    $$\left.\begin{matrix} n_{14y} \left(a_z - r_{14z}\right) - n_{14z} \left(a_y - r_{14y}\right) \\ - n_{14x} \left(a_z - r_{14z}\right) + n_{14z} \left(a_x - r_{14x}\right) \\ n_{14x} \left(a_y - r_{14y}\right) - n_{14y} \left(a_x - r_{14x}\right) \\ n_{14x} \\ n_{14y} \\ n_{14z} \end{matrix}\right.$$
    \\\qquad
    $$\left.\begin{matrix} - n_{35x} & n_{25y} \left(a_z - r_{25z}\right) - n_{25z} \left(a_y - r_{25y}\right) \\ - n_{35y} & - n_{25x} \left(a_z - r_{25z}\right) + n_{25z} \left(a_x - r_{25x}\right) \\ - n_{35z} & n_{25x} \left(a_y - r_{25y}\right) - n_{25y} \left(a_x - r_{25x}\right) \\ 0 & n_{25x} \\ 0 & n_{25y} \\ 0 & n_{25z} \end{matrix}\right.$$
    \\\qquad
    $$\left.\begin{matrix} - n_{35y} \left(a_z - r_{35z}\right) + n_{35z} \left(a_y - r_{35y}\right) & 0 \\ n_{35x} \left(a_z - r_{35z}\right) - n_{35z} \left(a_x - r_{35x}\right) & - a_z + r_{42z} \\ - n_{35x} \left(a_y - r_{35y}\right) + n_{35y} \left(a_x - r_{35x}\right) & a_y - r_{42y} \\ - n_{35x} & 1 \\ - n_{35y} & 0 \\ - n_{35z} & 0 \end{matrix}\right.$$
    \\\qquad
    $$\left.\begin{matrix} a_z - r_{42z} & - a_y + r_{42y}\\ 0 & a_x - r_{42x}\\ - a_x + r_{42x} & 0\\ 0 & 0\\ 1 & 0\\ 0 & 1\end{matrix}\right]
    \begin{Bmatrix}\dot{\theta}_{13}\\\dot{\theta}_{14}\\\dot{d}_{35}\\\dot{\theta}_{25}\\\dot{\theta}_{35}\\\omega_{42x}\\\omega_{42y}\\\omega_{42z}\end{Bmatrix}
    =
    \begin{Bmatrix}0\\0\\0\\0\\0\\0\end{Bmatrix}$$

    $$\Rightarrow
    \left[\begin{matrix}- n_{13y} \left(a_z - r_{13z}\right) + n_{13z} \left(a_y - r_{13y}\right) \\n_{13x} \left(a_z - r_{13z}\right) - n_{13z} \left(a_x - r_{13x}\right) \\- n_{13x} \left(a_y - r_{13y}\right) + n_{13y} \left(a_x - r_{13x}\right) \\- n_{13x} \\- n_{13y} \\- n_{13z} \end{matrix}\right.$$
    \\\qquad
    $$\left.\begin{matrix} n_{14y} \left(a_z - r_{14z}\right) - n_{14z} \left(a_y - r_{14y}\right)\\ - n_{14x} \left(a_z - r_{14z}\right) + n_{14z} \left(a_x - r_{14x}\right)\\ n_{14x} \left(a_y - r_{14y}\right) - n_{14y} \left(a_x - r_{14x}\right)\\ n_{14x}\\ n_{14y}\\ n_{14z}\end{matrix}\right]
    \begin{Bmatrix}\dot{\theta}_{13}\\\dot{\theta}_{14}\end{Bmatrix} +$$
    \\\qquad
    $$\left[\begin{matrix}- n_{35x} & n_{25y} \left(a_z - r_{25z}\right) - n_{25z} \left(a_y - r_{25y}\right) \\- n_{35y} & - n_{25x} \left(a_z - r_{25z}\right) + n_{25z} \left(a_x - r_{25x}\right) \\- n_{35z} & n_{25x} \left(a_y - r_{25y}\right) - n_{25y} \left(a_x - r_{25x}\right) \\0 & n_{25x} \\0 & n_{25y} \\0 & n_{25z} \end{matrix}\right.$$
    \\\qquad
    $$\left.\begin{matrix} - n_{35y} \left(a_z - r_{35z}\right) + n_{35z} \left(a_y - r_{35y}\right) & 0 \\ n_{35x} \left(a_z - r_{35z}\right) - n_{35z} \left(a_x - r_{35x}\right) & - a_z + r_{42z} \\ - n_{35x} \left(a_y - r_{35y}\right) + n_{35y} \left(a_x - r_{35x}\right) & a_y - r_{42y} \\ - n_{35x} & 1 \\ - n_{35y} & 0 \\ - n_{35z} & 0 \end{matrix}\right.$$
    \\\qquad
    $$\left.\begin{matrix} a_z - r_{42z} & - a_y + r_{42y}\\ 0 & a_x - r_{42x}\\ - a_x + r_{42x} & 0\\ 0 & 0\\ 1 & 0\\ 0 & 1\end{matrix}\right]
    \begin{Bmatrix}\dot{d}_{35}\\\dot{\theta}_{25}\\\dot{\theta}_{35}\\\omega_{42x}\\\omega_{42y}\\\omega_{42z}\end{Bmatrix}
    =
    \begin{Bmatrix}0\\0\\0\\0\\0\\0\end{Bmatrix}$$

\[\Rightarrow \left[A_1\right]\{\dot{\theta}_a\}+\left[A_2\right]\{\Omega\}=0\] \[\Rightarrow \{\Omega\}=-\left[A_2\right]^{-1}\left[A_1\right]\{\dot{\theta}_a\}\]\[\therefore \{\dot{X}\}=\left[J\right] \{\dot{\theta}\}=\left[J_1\right] \{\dot{\theta}_a\}+\left[J_2\right] \{\Omega\}\]\[=\left[J_1\right] \{\dot{\theta}_a\}-\left[J_2\right]\left[A_2\right]^{-1}\left[A_1\right] \{\dot{\theta}_a\}\]

The elements of $\hat{n}_{ij}=\begin{Bmatrix} n_{ijx} & n_{ijy} & n_{ijz} \end{Bmatrix}^T$ should always satisfy the equation $n_{ijx}^2+n_{ijy}^2+n_{ijz}^2=1$. For this to be satisfied, the elements are considered as below: \[\hat{n}_{ij}=\begin{Bmatrix} n_{ijx} \\ n_{ijy} \\ n_{ijz} \end{Bmatrix}=\begin{Bmatrix} \sin{\left(\beta_{ij}\right)} \cos{\left(\phi_{ij}\right)} \\ \sin{\left(\beta_{ij}\right)} \sin{\left(\phi_{ij}\right)} \\ \cos{\left(\beta_{ij}\right)} \end{Bmatrix}\]
For the chosen end-effector location $\vec{a}=\begin{Bmatrix} 3 & 4 & 5 \end{Bmatrix}^T$, the matrices $\left[J_1\right]$, $\left[J_2\right]$, $\left[A_1\right]$ and $\left[A_2\right]$ would finally be\newline

\resizebox{0.45\textwidth}{!}{$
$$\left[J_1\right]=\left[\begin{matrix}- \left(4 - r_{13y}\right) \cos{\left(\beta_{13} \right)} + \left(5 - r_{13z}\right) \sin{\left(\phi_{13} \right)} \sin{\left(\beta_{13} \right)} & 0\\\left(3 - r_{13x}\right) \cos{\left(\beta_{13} \right)} - \left(5 - r_{13z}\right) \sin{\left(\beta_{13} \right)} \cos{\left(\phi_{13} \right)} & 0\\- \left(3 - r_{13x}\right) \sin{\left(\phi_{13} \right)} \sin{\left(\beta_{13} \right)} + \left(4 - r_{13y}\right) \sin{\left(\beta_{13} \right)} \cos{\left(\phi_{13} \right)} & 0\\\sin{\left(\beta_{13} \right)} \cos{\left(\phi_{13} \right)} & 0\\\sin{\left(\phi_{13} \right)} \sin{\left(\beta_{13} \right)} & 0\\\cos{\left(\beta_{13} \right)} & 0\end{matrix}\right]$$
$}

\resizebox{0.2\textwidth}{!}{$
$$\left[J_2\right]=
\left[\begin{matrix}\sin{\left(\beta_{35} \right)} \cos{\left(\phi_{35} \right)} & 0 \\\sin{\left(\phi_{35} \right)} \sin{\left(\beta_{35} \right)} & 0 \\\cos{\left(\beta_{35} \right)} & 0 \\0 & 0 \\0 & 0 \\0 & 0 \end{matrix}\right.$$
$}
\\\qquad
\resizebox{0.4\textwidth}{!}{$
$$\left.\begin{matrix} - \left(4 - r_{35y}\right) \cos{\left(\beta_{35} \right)} + \left(5 - r_{35z}\right) \sin{\left(\phi_{35} \right)} \sin{\left(\beta_{35} \right)} \\ \left(3 - r_{35x}\right) \cos{\left(\beta_{35} \right)} - \left(5 - r_{35z}\right) \sin{\left(\beta_{35} \right)} \cos{\left(\phi_{35} \right)} \\ - \left(3 - r_{35x}\right) \sin{\left(\phi_{35} \right)} \sin{\left(\beta_{35} \right)} + \left(4 - r_{35y}\right) \sin{\left(\beta_{35} \right)} \cos{\left(\phi_{35} \right)} \\ \sin{\left(\beta_{35} \right)} \cos{\left(\phi_{35} \right)} \\ \sin{\left(\phi_{35} \right)} \sin{\left(\beta_{35} \right)} \\ \cos{\left(\beta_{35} \right)} \end{matrix}\right.$$
$}
\\\qquad
$$\resizebox{0.08\textwidth}{!}{$\left.\begin{matrix} 0 & 0 & 0\\ 0 & 0 & 0\\ 0 & 0 & 0\\ 0 & 0 & 0\\ 0 & 0 & 0\\ 0 & 0 & 0\end{matrix}\right]$}
$$

  \resizebox{0.45\textwidth}{!}{$
    $$\left[A_1\right]=\left[\begin{matrix}\left(4 - r_{13y}\right) \cos{\left(\beta_{13} \right)} - \left(5 - r_{13z}\right) \sin{\left(\phi_{13} \right)} \sin{\left(\beta_{13} \right)} \\- \left(3 - r_{13x}\right) \cos{\left(\beta_{13} \right)} + \left(5 - r_{13z}\right) \sin{\left(\beta_{13} \right)} \cos{\left(\phi_{13} \right)} \\\left(3 - r_{13x}\right) \sin{\left(\phi_{13} \right)} \sin{\left(\beta_{13} \right)} - \left(4 - r_{13y}\right) \sin{\left(\beta_{13} \right)} \cos{\left(\phi_{13} \right)} \\- \sin{\left(\beta_{13} \right)} \cos{\left(\phi_{13} \right)} \\- \sin{\left(\phi_{13} \right)} \sin{\left(\beta_{13} \right)} \\- \cos{\left(\beta_{13} \right)} \end{matrix}\right.$$
    $}
    \\\qquad
    \resizebox{0.45\textwidth}{!}{$
    $$\left.\begin{matrix} - \left(4 - r_{14y}\right) \cos{\left(\beta_{14} \right)} + \left(5 - r_{14z}\right) \sin{\left(\phi_{14} \right)} \sin{\left(\beta_{14} \right)}\\ \left(3 - r_{14x}\right) \cos{\left(\beta_{14} \right)} - \left(5 - r_{14z}\right) \sin{\left(\beta_{14} \right)} \cos{\left(\phi_{14} \right)}\\ - \left(3 - r_{14x}\right) \sin{\left(\phi_{14} \right)} \sin{\left(\beta_{14} \right)} + \left(4 - r_{14y}\right) \sin{\left(\beta_{14} \right)} \cos{\left(\phi_{14} \right)}\\ \sin{\left(\beta_{14} \right)} \cos{\left(\phi_{14} \right)}\\ \sin{\left(\phi_{14} \right)} \sin{\left(\beta_{14} \right)}\\ \cos{\left(\beta_{14} \right)}\end{matrix}\right]$$
    $}

    \resizebox{0.22\textwidth}{!}{$
    $$\left[A_2\right]=\left[\begin{matrix}- \sin{\left(\beta_{35} \right)} \cos{\left(\phi_{35} \right)} \\- \sin{\left(\phi_{35} \right)} \sin{\left(\beta_{35} \right)} \\- \cos{\left(\beta_{35} \right)} \\0 \\0 \\0 \end{matrix}\right.$$
    $}
    \\\qquad
    \resizebox{0.45\textwidth}{!}{$
    $$\left.\begin{matrix} - \left(4 - r_{25y}\right) \cos{\left(\beta_{25} \right)} + \left(5 - r_{25z}\right) \sin{\left(\phi_{25} \right)} \sin{\left(\beta_{25} \right)} \\ \left(3 - r_{25x}\right) \cos{\left(\beta_{25} \right)} - \left(5 - r_{25z}\right) \sin{\left(\beta_{25} \right)} \cos{\left(\phi_{25} \right)} \\ - \left(3 - r_{25x}\right) \sin{\left(\phi_{25} \right)} \sin{\left(\beta_{25} \right)} + \left(4 - r_{25y}\right) \sin{\left(\beta_{25} \right)} \cos{\left(\phi_{25} \right)} \\ \sin{\left(\beta_{25} \right)} \cos{\left(\phi_{25} \right)} \\ \sin{\left(\phi_{25} \right)} \sin{\left(\beta_{25} \right)} \\ \cos{\left(\beta_{25} \right)} \end{matrix}\right.$$
    $}
    \\\qquad
    \resizebox{0.45\textwidth}{!}{$
    $$\left.\begin{matrix} \left(4 - r_{35y}\right) \cos{\left(\beta_{35} \right)} - \left(5 - r_{35z}\right) \sin{\left(\phi_{35} \right)} \sin{\left(\beta_{35} \right)} \\ - \left(3 - r_{35x}\right) \cos{\left(\beta_{35} \right)} + \left(5 - r_{35z}\right) \sin{\left(\beta_{35} \right)} \cos{\left(\phi_{35} \right)} \\ \left(3 - r_{35x}\right) \sin{\left(\phi_{35} \right)} \sin{\left(\beta_{35} \right)} - \left(4 - r_{35y}\right) \sin{\left(\beta_{35} \right)} \cos{\left(\phi_{35} \right)} \\ - \sin{\left(\beta_{35} \right)} \cos{\left(\phi_{35} \right)} \\ - \sin{\left(\phi_{35} \right)} \sin{\left(\beta_{35} \right)} \\ - \cos{\left(\beta_{35} \right)} \end{matrix}\right.$$
    $}
    \\\qquad
    
    $$\left.\begin{matrix} 0 & 5 - r_{42z} & r_{42y} - 4\\ r_{42z} - 5 & 0 & 3 - r_{42x}\\ 4 - r_{42y} & r_{42x} - 3 & 0\\ 1 & 0 & 0\\ 0 & 1 & 0\\ 0 & 0 & 1\end{matrix}\right]$$
By using the matrices $\left[J_1\right]$, $\left[J_2\right]$, $\left[A_1\right]$ and $\left[A_2\right]$, the relation between the active joint velocities and the end-effector velocities can be written as

\[
   \{\dot{X}\}=\left( \left[J_1\right]-\left[J_2\right] \left[A_2\right]^{-1} \left[A_1\right] \right) \{\dot{\theta}_a\}=\left[\widetilde{J}\right] \{\dot{\theta}_a\} \]
   where $\left[\widetilde{J}\right]=\left[J_1\right]-\left[J_2\right] \left[A_2\right]^{-1} \left[A_1\right] $ is the Jacobian matrix of the manipulator for the considered actuating joints.
\newline

Therefore, the formulation of the optimisation problem would then be
\newline
Maximise\[f_1=\sqrt{\det{\left(\left[\widetilde{J}\right]^T \left[\widetilde{J}\right]\right)}}\]

\hl{subject} to the bounds
$$\begin{matrix}
  0 \leq r_{ijk} \leq 10, \\
  0 \leq \beta_{ij} \leq \pi, \\
  0 \leq \phi_{ij} \leq 2\pi.
  \end{matrix}
$$
Based on the strategy mentioned in section \ref{strategy_used_in_optimisation}, the above optimisation problem is attempted for solution, but it did not pass the criterion mentioned in Step 1. Hence, the modified objective function is taken for optimisation in order to avoid type-2 singularity. The modified optimisation problem would then become
\newline
Maximise\[f_2=\sqrt{\det{\left(\left[\widetilde{A}\right]^T \left[\widetilde{A}\right]\right)} \times \det{\left(\left[\widetilde{B}\right]^T \left[\widetilde{B}\right]\right)}}\]

\hl{subject} to the bounds
$$\begin{matrix}
  0 \leq r_{ijk} \leq 10, \\
  0 \leq \beta_{ij} \leq \pi, \\
  0 \leq \phi_{ij} \leq 2\pi,
  \end{matrix}
$$

\noindent\emph{Condition number formulation:}

Scaled Jacobian matrix is formulated based on the methodology explained in section \ref{methodology_of_scaled_matrix}, as shown below.

The effective distances are formulated as follows.

$$\begin{matrix}
  \bar{d}_{1}=\left|\left(\vec{r}_{13}-\vec{a}\right)\times \hat{n}_{13}\right| \\
  \bar{d}_{2}=\left|\left(\vec{r}_{14}-\vec{a}\right)\times \hat{n}_{14}\right| \\
  \bar{d}_{3}=\left|\left(\vec{r}_{25}-\vec{a}\right)\times \hat{n}_{25}\right| \\
  \bar{d}_{4}=\left|\left(\vec{r}_{35}-\vec{a}\right)\times \hat{n}_{35}\right| \\
  \bar{d}_{5}=\left|\vec{r}_{42}-\vec{a}\right|
\end{matrix}
$$

The characteristic length is formulated as follows.
\[L=\frac{\bar{d}_{1}+\bar{d}_{2}+\bar{d}_{3}+\bar{d}_{4}+\bar{d}_{5}}{5}\]

By using the above characteristic length, the scaling matrix $\left[S\right]$ can be calculated as follows.
\[\left[S\right]=\begin{bmatrix}
  \frac{1}{L} & 0 & 0 & 0 & 0 & 0\\
  0 & \frac{1}{L} & 0 & 0 & 0 & 0\\
  0 & 0 & \frac{1}{L} & 0 & 0 & 0\\
  0 & 0 & 0 & 1 & 0 & 0\\
  0 & 0 & 0 & 0 & 1 & 0\\
  0 & 0 & 0 & 0 & 0 & 1\\
  \end{bmatrix}\]
  
The scaled Jacobian matrix can then be calculated as
\[\left[\widetilde{J}_s\right]=\left[S\right]\left[\widetilde{J}\right]\]
The condition number of $\left[\widetilde{J}_s\right]$ is used for comparison.

\noindent\emph{Result of the optimisation problem:}

Solution to the optimisation problem is shown in Table \ref{Tab:table_5_71} (with the values rounded up to two decimal places).
\begin{table}[h]  \centering \begin{tabular}{|c|c|}
\hline
Variable & Value \\ 
\hline
$\phi_{13}$ & $\ang{360.0}$ \\
\hline$\phi_{14}$ & $\ang{104.74}$ \\
\hline$\phi_{25}$ & $\ang{347.83}$ \\
\hline$\phi_{35}$ & $\ang{32.44}$ \\
\hline$r_{13x}$ & $10.0$ \\
\hline$r_{13y}$ & $10.0$ \\
\hline$r_{13z}$ & $10.0$ \\
\hline$r_{14x}$ & $10.0$ \\
\hline$r_{14y}$ & $10.0$ \\
\hline$r_{14z}$ & $0.0$ \\
\hline$r_{42x}$ & $10.0$ \\
\hline$r_{42y}$ & $0.0$ \\
\hline$r_{42z}$ & $10.0$ \\
\hline$r_{25x}$ & $0.0$ \\
\hline$r_{25y}$ & $0.0$ \\
\hline$r_{25z}$ & $0.0$ \\
\hline$r_{35x}$ & $0.0$ \\
\hline$r_{35y}$ & $10.0$ \\
\hline$r_{35z}$ & $0.0$ \\
\hline$\beta_{13}$ & $\ang{137.39}$ \\
\hline$\beta_{14}$ & $\ang{45.96}$ \\
\hline$\beta_{25}$ & $\ang{134.35}$ \\
\hline$\beta_{35}$ & $\ang{68.46}$ \\
\hline\end{tabular} \caption{Result of the optimisation problem corresponding to 2D-M71 manipulator.} \label{Tab:table_5_71} \end{table}
\newline
Therefore, the locations and orientations of the joints are shown below.
\newline
$$\begin{matrix}
  \hat{n}_{13}=-0.0\hat{i}+0.68\hat{j}-0.74\hat{k} \\
  \hat{n}_{14}=0.7\hat{i}-0.18\hat{j}+0.7\hat{k} \\
  \hat{n}_{25}=-0.15\hat{i}+0.7\hat{j}-0.7\hat{k} \\
  \hat{n}_{35}=0.5\hat{i}+0.78\hat{j}+0.37\hat{k} \\
  \vec{r}_{13}=10.0\hat{i}+10.0\hat{j}+10.0\hat{k} \\
  \vec{r}_{14}=10.0\hat{i}+10.0\hat{j}+0.0\hat{k} \\
  \vec{r}_{42}=10.0\hat{i}+0.0\hat{j}+10.0\hat{k} \\
  \vec{r}_{25}=0.0\hat{i}+0.0\hat{j}+0.0\hat{k} \\
  \vec{r}_{35}=0.0\hat{i}+10.0\hat{j}+0.0\hat{k}
\end{matrix}
$$

The corresponding optimum manipulability is given by \[\mu=173.16\]

And the corresponding condition number is \[\kappa=1.98\]

And the corresponding scaled manipulability index, which is the product of singular values of the scaled Jacobian $\left[\widetilde{J}_s\right]$, is \[\bar{\mu}=0.2305\]

%% file: contentfolder/mechanisms2/1dof_manipulabilitytable_final.tex
\small
\begin{center}

\tablefirsthead{%

\hline
\multicolumn{1}{|c|}{S.No.} &
Name &
$\bar{\mu}$ &
$\mu$ &
\multicolumn{1}{c|}{$f$} &
AJV \\
\hline}
\tablehead{%
\hline
\multicolumn{6}{|l|}{\small\sl continuing...}\\
\hline
\multicolumn{1}{|c|}{S.No.} &
Name &
$\bar{\mu}$ &
$\mu$ &
\multicolumn{1}{c|}{$f$} &
AJV \\
\hline}
\tabletail{%
\hline
\multicolumn{6}{|r|}{\small\sl ...continued}\\
\hline}
\tablelasttail{\hline}
\label{Tab:table_dof1mt_1}
\bottomcaption{Prescription of 1-DOF Manipulators.}


\begin{supertabular}{|r|r|r|r|r|r|}
\hline
1 & 1D-M10 & 0.1523 & 8.1 & $f_1$ & $\dot{\theta}_{14}$ \\
\hline2 & 1D-M65 & 0.1376 & 10.5 & $f_2$ & $\dot{\theta}_{14}$ \\
\hline3 & 1D-M8 & 0.1152 & 7.1 & $f_1$ & $\dot{\theta}_{14}$ \\
\hline4 & 1D-M94 & 0.1106 & 1.0 & $f_2$ & $\dot{d}_{24}$ \\
\hline5 & 1D-M28 & 0.1088 & 8.1 & $f_1$ & $\dot{\theta}_{14}$ \\
\hline6 & 1D-M20 & 0.0824 & 10.5 & $f_1$ & $\dot{\theta}_{14}$ \\
\hline7 & 1D-M34 & 0.0798 & 7.1 & $f_1$ & $\dot{\theta}_{14}$ \\
\hline8 & 1D-M95 & 0.0776 & 1.0 & $f_2$ & $\dot{d}_{12}$ \\
\hline9 & 1D-M7 & 0.0774 & 8.4 & $f_1$ & $\dot{\theta}_{14}$ \\
\hline10 & 1D-M29 & 0.0732 & 10.5 & $f_1$ & $\dot{\theta}_{14}$ \\
\hline11 & 1D-M41 & 0.0722 & 9.5 & $f_1$ & $\dot{\theta}_{14}$ \\
\hline12 & 1D-M40 & 0.071 & 7.1 & $f_1$ & $\dot{\theta}_{14}$ \\
\hline13 & 1D-M35 & 0.0706 & 10.5 & $f_1$ & $\dot{\theta}_{14}$ \\
\hline14 & 1D-M11 & 0.062 & 10.5 & $f_1$ & $\dot{\theta}_{14}$ \\
\hline15 & 1D-M64 & 0.0563 & 20.3 & $f_2$ & $\dot{\theta}_{13}$ \\
\hline16 & 1D-M66 & 0.0487 & 19.8 & $f_2$ & $\dot{\theta}_{23}$ \\
\hline17 & 1D-M60 & 0.0475 & 8.4 & $f_1$ & $\dot{\theta}_{14}$ \\
\hline18 & 1D-M93 & 0.0467 & 1.0 & $f_2$ & $\dot{d}_{23}$ \\
\hline19 & 1D-M59 & 0.0462 & 9.5 & $f_1$ & $\dot{\theta}_{14}$ \\
\hline20 & 1D-M91 & 0.0438 & 1.0 & $f_2$ & $\dot{d}_{13}$ \\
\hline21 & 1D-M79 & 0.0335 & 1 & $f_1$ & $\dot{d}_{14}$ \\
\hline22 & 1D-M25 & 0.0313 & 17.3 & $f_2$ & $\dot{\theta}_{13}$ \\
\hline23 & 1D-M44 & 0.0313 & 17.3 & $f_2$ & $\dot{\theta}_{23}$ \\
\hline24 & 1D-M67 & 0.0306 & 18.3 & $f_2$ & $\dot{\theta}_{24}$ \\
\hline25 & 1D-M15 & 0.0305 & 21.2 & $f_2$ & $\dot{\theta}_{12}$ \\
\hline26 & 1D-M22 & 0.0246 & 17.3 & $f_2$ & $\dot{\theta}_{13}$ \\
\hline27 & 1D-M13 & 0.024 & 15.3 & $f_2$ & $\dot{\theta}_{12}$ \\
\hline28 & 1D-M46 & 0.0237 & 15.3 & $f_2$ & $\dot{\theta}_{12}$ \\
\hline29 & 1D-M45 & 0.0229 & 17.3 & $f_2$ & $\dot{\theta}_{24}$ \\
\hline30 & 1D-M3 & 0.0222 & 15.4 & $f_2$ & $\dot{\theta}_{13}$ \\
\hline31 & 1D-M49 & 0.0219 & 14.1 & $f_2$ & $\dot{\theta}_{12}$ \\
\hline32 & 1D-M68 & 0.0214 & 19.6 & $f_2$ & $\dot{\theta}_{12}$ \\
\hline33 & 1D-M5 & 0.0212 & 14.7 & $f_2$ & $\dot{\theta}_{13}$ \\
\hline34 & 1D-M47 & 0.0212 & 17.3 & $f_2$ & $\dot{\theta}_{12}$ \\
\hline35 & 1D-M24 & 0.0212 & 17.3 & $f_2$ & $\dot{\theta}_{13}$ \\
\hline36 & 1D-M51 & 0.0203 & 14.9 & $f_2$ & $\dot{\theta}_{12}$ \\
\hline37 & 1D-M18 & 0.0194 & 14.6 & $f_2$ & $\dot{\theta}_{24}$ \\
\hline38 & 1D-M69 & 0.019 & 19.0 & $f_2$ & $\dot{\theta}_{24}$ \\
\hline39 & 1D-M48 & 0.0189 & 14.0 & $f_2$ & $\dot{\theta}_{12}$ \\
\hline40 & 1D-M12 & 0.0188 & 12.8 & $f_2$ & $\dot{\theta}_{23}$ \\
\hline41 & 1D-M9 & 0.0185 & 12.8 & $f_2$ & $\dot{\theta}_{23}$ \\
\hline42 & 1D-M23 & 0.0177 & 13.0 & $f_2$ & $\dot{\theta}_{13}$ \\
\hline43 & 1D-M62 & 0.0167 & 9.6 & $f_2$ & $\dot{\theta}_{12}$ \\
\hline44 & 1D-M26 & 0.0167 & 9.6 & $f_2$ & $\dot{\theta}_{13}$ \\
\hline45 & 1D-M14 & 0.0163 & 10.1 & $f_2$ & $\dot{\theta}_{12}$ \\
\hline46 & 1D-M96 & 0.016 & 1.0 & $f_2$ & $\dot{d}_{24}$ \\
\hline47 & 1D-M6 & 0.0159 & 12.0 & $f_2$ & $\dot{\theta}_{13}$ \\
\hline48 & 1D-M4 & 0.0158 & 13.4 & $f_2$ & $\dot{\theta}_{13}$ \\
\hline49 & 1D-M89 & 0.0155 & 1 & $f_1$ & $\dot{d}_{14}$ \\
\hline50 & 1D-M38 & 0.0153 & 12.9 & $f_2$ & $\dot{\theta}_{23}$ \\
\hline51 & 1D-M58 & 0.0148 & 8.2 & $f_2$ & $\dot{\theta}_{13}$ \\
\hline52 & 1D-M57 & 0.0143 & 11.7 & $f_2$ & $\dot{\theta}_{34}$ \\
\hline53 & 1D-M76 & 0.0138 & 1 & $f_1$ & $\dot{d}_{14}$ \\
\hline54 & 1D-M17 & 0.0136 & 8.6 & $f_2$ & $\dot{\theta}_{12}$ \\
\hline55 & 1D-M50 & 0.0135 & 11.1 & $f_2$ & $\dot{\theta}_{12}$ \\
\hline56 & 1D-M19 & 0.0134 & 10.3 & $f_2$ & $\dot{\theta}_{13}$ \\
\hline57 & 1D-M56 & 0.0129 & 10.5 & $f_2$ & $\dot{\theta}_{24}$ \\
\hline58 & 1D-M21 & 0.0128 & 10.5 & $f_2$ & $\dot{\theta}_{12}$ \\
\hline59 & 1D-M27 & 0.0127 & 9.3 & $f_2$ & $\dot{\theta}_{13}$ \\
\hline60 & 1D-M77 & 0.0126 & 1 & $f_1$ & $\dot{d}_{14}$ \\
\hline61 & 1D-M16 & 0.0125 & 9.3 & $f_2$ & $\dot{\theta}_{12}$ \\
\hline62 & 1D-M2 & 0.0116 & 6.7 & $f_2$ & $\dot{\theta}_{13}$ \\
\hline63 & 1D-M1 & 0.0113 & 8.4 & $f_2$ & $\dot{\theta}_{13}$ \\
\hline64 & 1D-M43 & 0.0108 & 1 & $f_1$ & $\dot{d}_{14}$ \\
\hline65 & 1D-M39 & 0.01 & 8.5 & $f_2$ & $\dot{\theta}_{24}$ \\
\hline66 & 1D-M80 & 0.0082 & 1 & $f_1$ & $\dot{d}_{14}$ \\
\hline67 & 1D-M36 & 0.0081 & 1 & $f_1$ & $\dot{d}_{14}$ \\
\hline68 & 1D-M92 & 0.0071 & 1 & $f_1$ & $\dot{d}_{14}$ \\
\hline69 & 1D-M61 & 0.0051 & 1 & $f_1$ & $\dot{d}_{14}$ \\
\hline70 & 1D-M42 & 0.0034 & 1 & $f_1$ & $\dot{d}_{14}$ \\
\hline71 & 1D-M31 & 0.0031 & 1.7 & $f_2$ & $\dot{d}_{13}$ \\
\hline72 & 1D-M37 & 0.0025 & 1 & $f_1$ & $\dot{d}_{14}$ \\
\hline73 & 1D-M86 & 0.0021 & 1.0 & $f_2$ & $\dot{d}_{12}$ \\
\hline74 & 1D-M30 & 0.002 & 1.1 & $f_2$ & $\dot{d}_{13}$ \\
\hline75 & 1D-M53 & 0.0017 & 1.1 & $f_2$ & $\dot{d}_{12}$ \\
\hline76 & 1D-M83 & 0.0016 & 1.0 & $f_2$ & $\dot{d}_{12}$ \\
\hline77 & 1D-M84 & 0.0015 & 1.0 & $f_2$ & $\dot{d}_{12}$ \\
\hline78 & 1D-M87 & 0.0015 & 1.0 & $f_2$ & $\dot{d}_{24}$ \\
\hline79 & 1D-M55 & 0.0015 & 1.3 & $f_2$ & $\dot{d}_{12}$ \\
\hline80 & 1D-M71 & 0.0015 & 1.0 & $f_2$ & $\dot{d}_{13}$ \\
\hline81 & 1D-M73 & 0.0015 & 1.0 & $f_2$ & $\dot{d}_{13}$ \\
\hline82 & 1D-M90 & 0.0015 & 1.0 & $f_2$ & $\dot{d}_{12}$ \\
\hline83 & 1D-M52 & 0.0015 & 1.1 & $f_2$ & $\dot{d}_{12}$ \\
\hline84 & 1D-M82 & 0.0014 & 1.0 & $f_2$ & $\dot{d}_{12}$ \\
\hline85 & 1D-M72 & 0.0014 & 1 & $f_2$ & $\dot{d}_{13}$ \\
\hline86 & 1D-M70 & 0.0014 & 1.0 & $f_2$ & $\dot{d}_{13}$ \\
\hline87 & 1D-M81 & 0.0014 & 1 & $f_2$ & $\dot{d}_{23}$ \\
\hline88 & 1D-M85 & 0.0014 & 1.0 & $f_2$ & $\dot{d}_{12}$ \\
\hline89 & 1D-M88 & 0.0014 & 1.0 & $f_2$ & $\dot{d}_{13}$ \\
\hline90 & 1D-M63 & 0.0013 & 1.1 & $f_2$ & $\dot{d}_{12}$ \\
\hline91 & 1D-M32 & 0.0013 & 0.7 & $f_2$ & $\dot{d}_{13}$ \\
\hline92 & 1D-M54 & 0.0013 & 1.1 & $f_2$ & $\dot{d}_{12}$ \\
\hline93 & 1D-M33 & 0.0012 & 0.8 & $f_2$ & $\dot{d}_{13}$ \\
\hline94 & 1D-M78 & 0.0009 & 0.6 & $f_2$ & $\dot{d}_{23}$ \\
\hline95 & 1D-M74 & 0.0008 & 0.6 & $f_2$ & $\dot{d}_{13}$ \\
\hline96 & 1D-M75 & 0.0006 & 0.4 & $f_2$ & $\dot{d}_{13}$ \\
\hline

\end{supertabular}
\end{center}


\normalsize

%% file: contentfolder/mechanisms2/manipulabilitytable_new3.tex
\begin{table}[h]  \centering \begin{tabular}{|c|c|c|c|c|c|c|}
\hline
S.No. & Name & $\bar{\mu}$ & $\kappa$ & $i_{\kappa}$ & $\mu$ & $f$ \\ 
\hline
1 & 3D-M1 & 1.0113 & 1.5 & 2 & 1058.9 & $f_1$ \\
\hline2 & 3D-M8 & 1.0 & 1.0 & 1 & 1.0 & $f_1$ \\
\hline3 & 3D-M2 & 0.0962 & 14.8 & 4 & 111.0 & $f_1$ \\
\hline4 & 3D-M3 & 0.0962 & 15.5 & 8 & 111.0 & $f_1$ \\
\hline5 & 3D-M4 & 0.0962 & 14.8 & 7 & 111.0 & $f_1$ \\
\hline6 & 3D-M7 & 0.0091 & 14.8 & 5 & 10.5 & $f_1$ \\
\hline7 & 3D-M5 & 0.0091 & 14.8 & 6 & 10.5 & $f_1$ \\
\hline8 & 3D-M6 & 0.0091 & 14.8 & 3 & 10.5 & $f_1$ \\
\hline\end{tabular} \caption{Prescription of 3-DOF Manipulators.} \label{Tab:table_dof3mt_1} \end{table}

%% file: contentfolder/mechanisms2/manipulabilitytable_new4.tex
\begin{table}[h]  \centering \begin{tabular}{|c|c|c|c|c|c|c|}
\hline
S.No. & Name & $\bar{\mu}$ & $\kappa$ & $i_{\kappa}$ & $\mu$ & $f$ \\ 
\hline
1 & 4D-M1 & 2.5607 & 1.5 & 1 & 2074.5 & $f_1$ \\
\hline2 & 4D-M13 & 0.4866 & 3.0 & 2 & 1.0 & $f_1$ \\
\hline3 & 4D-M12 & 0.4048 & 3.1 & 3 & 1 & $f_1$ \\
\hline4 & 4D-M4 & 0.2619 & 15.3 & 14 & 211.0 & $f_1$ \\
\hline5 & 4D-M5 & 0.2619 & 15.3 & 13 & 211.0 & $f_1$ \\
\hline6 & 4D-M2 & 0.2619 & 15.3 & 12 & 211.0 & $f_1$ \\
\hline7 & 4D-M3 & 0.2619 & 15.3 & 15 & 211.0 & $f_1$ \\
\hline8 & 4D-M15 & 0.077 & 4.9 & 4 & 1.0 & $f_1$ \\
\hline9 & 4D-M14 & 0.0538 & 5.5 & 5 & 1.0 & $f_1$ \\
\hline10 & 4D-M9 & 0.0253 & 12.9 & 7 & 17.6 & $f_1$ \\
\hline11 & 4D-M11 & 0.0253 & 12.9 & 10 & 17.6 & $f_1$ \\
\hline12 & 4D-M10 & 0.0253 & 12.9 & 6 & 17.6 & $f_1$ \\
\hline13 & 4D-M7 & 0.0253 & 12.9 & 8 & 17.6 & $f_1$ \\
\hline14 & 4D-M8 & 0.0253 & 12.9 & 9 & 17.6 & $f_1$ \\
\hline15 & 4D-M6 & 0.0253 & 12.9 & 11 & 17.6 & $f_1$ \\
\hline\end{tabular} \caption{Prescription of 4-DOF Manipulators.} \label{Tab:table_dof4mt_1} \end{table}

%% file: Appendices/Appendix_main.tex
\section*{Appendix A}
\label{AppendixA}

\input{contentfolder/mechanisms2/2dof_manipulabilitytable_final}

%% file: contentfolder/mechanisms2/2dof_manipulabilitytable_final.tex

\tiny
\begin{center}
\tablefirsthead{%
\hline
\multicolumn{1}{|c|}{S.No.} &
Name &
$\bar{\mu}$ &
$\kappa$ &
$i_{\kappa}$ &
$\mu$ &
\multicolumn{1}{c|}{$f$} &
AJV \\
\hline}
\tablehead{%
\hline
\multicolumn{8}{|l|}{\tiny\sl continuing...}\\
\hline
\multicolumn{1}{|c|}{S.No.} &
Name &
$\bar{\mu}$ &
$\kappa$ &
$i_{\kappa}$ &
$\mu$ &
\multicolumn{1}{c|}{$f$} &
AJV \\
\hline}
\tabletail{%
\hline
\multicolumn{8}{|r|}{\tiny\sl ...continued}\\
\hline}
\tablelasttail{\hline}
\bottomcaption{Prescription of 2-DOF Manipulators.}



\end{center}


\normalsize